\documentclass{article} % For LaTeX2e
\usepackage[preprint]{neurips_2025}

\usepackage{amsmath,amsfonts,bm}

\def\eqref#1{equation~\ref{#1}}
\def\1{\bm{1}}

\DeclareMathAlphabet{\mathsfit}{\encodingdefault}{\sfdefault}{m}{sl}
\SetMathAlphabet{\mathsfit}{bold}{\encodingdefault}{\sfdefault}{bx}{n}

\usepackage{caption}
\usepackage{subcaption}

\usepackage[framemethod=tikz]{mdframed}
\usepackage[most]{tcolorbox}
\usepackage[normalem]{ulem}
\usepackage{algorithm}
\usepackage{algpseudocode}
\usepackage{amsfonts}
\usepackage{amsmath}
\usepackage{arydshln}
\usepackage{balance}
\usepackage{blindtext}
\usepackage{booktabs}
\usepackage{cancel}
\usepackage{caption}[font={footnotesize}]
\usepackage{color,soul}
\usepackage{comment}
\usepackage{enumitem}
\usepackage{fontawesome}
\usepackage{graphicx}
\usepackage{hyperref,xcolor}% http://ctan.org/pkg/{hyperref,xcolor}
\usepackage{hyperref}
\hypersetup{hidelinks}
\usepackage{listings}
\usepackage{mathtools}
\usepackage{multirow}
\usepackage{pifont}
\usepackage{tabularx}
\usepackage{threeparttable}
\usepackage{tikz}
\usepackage{xcolor,colortbl}
\usepackage{xspace}
\usepackage{wasysym}

\usepackage{makecell}
\usepackage{svg}

\definecolor{winered}{rgb}{0.7,0,0}
\definecolor{gray}{gray}{0.7}
\definecolor{darkpastelgreen}{rgb}{0.01, 0.75, 0.24}
\definecolor{cadmiumgreen}{rgb}{0.0, 0.42, 0.24}
\definecolor{brickred}{rgb}{0.8, 0.25, 0.33}
\definecolor{cornellred}{rgb}{0.7, 0.11, 0.11}
\definecolor{burgundy}{rgb}{0.5, 0.0, 0.13}
\definecolor{frenchblue}{rgb}{0.0, 0.45, 0.73}
\definecolor{light-gray}{gray}{0.92}
\definecolor{lightlight-gray}{gray}{0.97}
\definecolor{codegray}{gray}{0.90}
\definecolor{inputgray}{gray}{0.90}
\definecolor{darkgreen}{RGB}{40,125,40}

\hypersetup
{
colorlinks=true,
linkcolor=winered,
urlcolor={winered},
filecolor={winered},
citecolor={winered},
allcolors={winered}
}

\newcommand{\cmt}[1]{}

\newcommand{\homedir}{\raise.17ex\hbox{$\scriptstyle\sim$}}

\newcommand{\etal}{et al.\xspace}

\global\mdfdefinestyle{rtboxstyle}{%
linecolor=black,%
leftmargin=0cm,rightmargin=0cm,linewidth=0.5pt,
roundcorner=3,
skipbelow=0pt,backgroundcolor=lightlight-gray
}

\newcommand{\code}[1]{\texttt{#1}}

\mdfdefinestyle{MyFrame}{%
     roundcorner=1pt,
     innertopmargin=6pt,
     innerbottommargin=6pt,
     innerrightmargin=6pt,
     innerleftmargin=6pt,
    }

\graphicspath{{figure/}{figures/}}
\DeclareGraphicsExtensions{.pdf,.eps,.png,.jpg,.jpeg}

\long\def\comment#1{}

\renewcommand{\paragraph}[1]{\smallskip\noindent\emph{#1}\quad}
\def\Snospace~{\S{}}

\newcommand{\heading}[1]{{\vspace{2pt}\noindent\bf{#1}}} % inside section
\newcommand{\eg}{{\it e.g.,~}}
\newcommand{\ie}{{\it i.e.,~}}
\newcommand{\etc}{{\it etc.~}}

\newcommand{\nm}[1]{{\it (#1)\xspace}} % at beginning of section -- no extra space

\newcommand{\optional}[1]{}

\def\pname{{\textsc{ProvCreator}}\xspace}
\usepackage{fp}
\usepackage{svg}
\usepackage{xcolor}
\usepackage{booktabs}
\newif\ifcomments
\commentstrue

\ifcomments
\usepackage{todonotes}
	\newcommand{\kjee}[1]{\noindent\textcolor{red}{[KJEE: #1]}}
	\newcommand{\wc}[1]{}
	\newcommand{\jdw}[1]{\textcolor{green}{[{\bf JDW:} #1]}}
	\newcommand{\kunal}[1]{\textcolor{blue}{[{\bf KUNAL:} #1]}}
	\newcommand{\srk}[1]{\textcolor{lime}{[{\bf SRK:} #1]}}
	\newcommand{\txw}[1]{\textcolor{purple}{[{\bf TXW:} #1]}}
	\newcommand{\corr}[2]{\sout{#1} \hl{#2}}
\else
	\newcommand{\kjee}[1]{}
	\newcommand{\wc}[1]{}
	\newcommand{\jdw}[1]{}
	\newcommand{\kunal}[1]{}
	\newcommand{\srk}[1]{}
	\newcommand{\txw}[1]{}
	\newcommand{\corr}[2]{}
\fi

\def\inhouse{real\xspace}%
\def\gdss{GDSS\xspace}%
\usepackage[acronym]{glossaries}

\newacronym{hdl}{HDL}{High-level Dynamic Language}

\newacronym{ml}{ML}{Machine Learning}
\newcommand{\ml}{\gls*{ml}\xspace}

\newacronym{ai}{AI}{Artificial Intelligence}
\newcommand{\ai}{\gls*{ai}\xspace}

\newacronym{nn}{NN}{Neural Network}

\newacronym{gnn}{GNN}{Graph Neural Network}
\newcommand{\gnn}{\gls*{gnn}\xspace}

\newacronym{dnn}{DNN}{Deep Neural Network}

\newacronym{rnn}{RNN}{Recurrent Neural Network}

\newacronym{llm}{LLM}{Large Language Model}
\newcommand{\llm}{\gls*{llm}\xspace}
\newcommand{\llms}{\glspl*{llm}\xspace}

\newacronym{mlm}{MLM}{Masked Language Model}

\newacronym{gan}{GAN}{Generative Adversarial Network}

\newacronym{nlp}{NLP}{Natural Language Processing}

\newacronym{pl}{PL}{Programming Language}

\newacronym{bert}{BERT}{Bidirectional Encoder Representations from Transformers}

\newacronym{csn}{CSN}{Code Search Net}

\newacronym{sota}{SOTA}{State-Of-The-Art}

\newacronym{pypi}{PyPI}{Python Package Index}

\newacronym{cdm}{CDM}{Common Data Model}

\newacronym{tc}{TC}{Transparent Computing}

\newacronym{dt}{DT}{Decision Tree}

\newacronym{apt}{APT}{Advanced Persistent Threat}

\newacronym{cve}{CVE}{Common Vulnerabilities and Exposures}

\newacronym{ids}{IDS}{Intrusion Detection System}

\newacronym{av}{AV}{Anti-Virus}

\newacronym{edr}{EDR}{Endpoint Detection and Response}

\newacronym{poi}{POI}{Point-Of-Interest}
\newcommand{\poi}{\gls*{poi}\xspace}

\newacronym{vlan}{VLAN}{Virtual Local Area Network}

\newacronym{ctwo}{C2}{Command and Control}

\newacronym{csg}{CSG}{Computer Security Group}

\newacronym{wicys}{WiCyS}{Women In Cyber Security}

\newacronym{nsa}{NSA}{National Security Agency}

\newacronym{dod}{DoD}{Department of Defense}

\newacronym{lsm}{LSM}{Log Structured Merge}

\newacronym{me}{ME}{Microelectronics}

\newacronym{ci}{CI}{CyberInfrastructure}

\newacronym{soc}{SoC}{System-on-Chip}

\newacronym{cots}{COTS}{Custom-Off-The-Shelves}

\newacronym{tle}{TLE}{Two-line element set}

\newacronym{wal}{WAL}{Write Ahead Log}

\newacronym{obc}{OBC}{On-Board Computer}

\newacronym{dos}{DoS}{Denial-of-Services}

\newacronym{ttp}{TTP}{Tactics, techniques, and Procedures}

\newacronym{cti}{CTI}{Cyber Threat Intelligence}

\newacronym{ioc}{IOC}{Indicators Of Compromise}

\newacronym{jspoc}{JSpOC}{The Joint Space Operations Center}

\newacronym{cfg}{CFG}{Control Flow Graph}

\newacronym{cdg}{CDG}{Control Dependency Graph}

\newacronym{geo}{GEO}{Geo Stational Orbit}

\newacronym{seu}{SEU}{Single-Event-Upset}

\newacronym{cpg}{CPG}{Control Property Graph}

\newacronym{ooc}{OOC}{Out-Of-Core}

\newacronym{KV}{KV}{Key Value}

\usepackage[utf8]{inputenc} % allow utf-8 input
\usepackage[T1]{fontenc}    % use 8-bit T1 fonts
\usepackage{hyperref}       % hyperlinks
\usepackage{url}            % simple URL typesetting
\usepackage{booktabs}       % professional-quality tables
\usepackage{amsfonts}       % blackboard math symbols
\usepackage{nicefrac}       % compact symbols for 1/2, etc.
\usepackage{microtype}      % microtypography
\usepackage{xcolor}         % colors
\usepackage{wrapfig}

\title{\pname: Synthesizing Complex Heterogenous Graphs with Node and Edge Attributes}

\author{
    Tianhao Wang  \\
    The University of Texas at Dallas  \\
    \And
    Simon Klancher  \\
    The University of Texas at Dallas  \\
    \And
    Kunal Mukherjee  \\
    The University of Texas at Dallas  \\
    \And
    Josh Wiedemeier  \\
    The University of Texas at Dallas  \\
    \And
    Feng Chen  \\
    The University of Texas at Dallas  \\
    \And
    Murat Kantarcioglu  \\
    Virginia Tech  \\
    \And
    Kangkook Jee  \\
    The University of Texas at Dallas  \\
}

\begin{document}

\maketitle

%! root=../main.tex
\begin{abstract}
The rise of graph-structured data has driven interest in graph learning and synthetic data generation. While successful in text and image domains, synthetic graph generation remains challenging --- especially for real-world graphs with complex, heterogeneous schemas. Existing research has focused mostly on homogeneous structures with simple attributes, limiting their usefulness and relevance for application domains requiring semantic fidelity.

In this research, we introduce \pname, a synthetic graph framework designed for complex heterogeneous graphs with high-dimensional node and edge attributes. \pname formulates graph synthesis as a sequence generation task, enabling the use of transformer-based large language models. It features a versatile graph-to-sequence encoder-decoder that \nm{1} losslessly encodes graph structure and attributes, \nm{2} efficiently compresses large graphs for contextual modeling, and \nm{3} supports end-to-end, learnable graph generation.

To validate our research, we evaluate \pname on two challenging domains: \emph{system provenance graphs in cybersecurity} and \emph{knowledge graphs from IntelliGraph Benchmark Dataset}. In both cases, \pname captures intricate dependencies between structure and semantics, enabling the generation of realistic and privacy-aware synthetic datasets.
% We release our framework and datasets to support broader research in secure, scalable graph learning.
\end{abstract}

\section{Introduction}\label{sec:introduction}
Recent advancements in \ml and \ai research have increasingly focused on solving problems set in graph-structured datasets.
The rich expressivity and flexibility of graph-structured data has attracted substantial effort towards developing graph learning and processing techniques.
Real-world problems are naturally expressible as graphs that capture relationships between entities: social graphs capture interpersonal relationships, knowledge graphs encode the relationships between concepts, causal graphs capture dependencies among events, \etc \citep{snapnets}

The complexity of graph schemas ranges all from simple classical graphs with pure nodes and undirected edges all the way to modern heterogeneous graphs where every node and edge contains arbitrary attributes of varying data types.
The unique combination of practicality and elegant theoretical backing makes graph-based \ml an attractive research area.

However, some graph datasets are restricted in scope or expensive to collect, which limits the data available to train advanced graph models.
%MK-05-14-25: Please put few citations to the next sentence to justify the claim.
Following the success of supplementing training data with \emph{synthetic data} in the text and image domains~\citep{xu2024surveyknowledgedistillationlarge,nvidia2025cosmosworldfoundationmodel}, the generation of usable \emph{synthetic graph data} has emerged as a prominent research topic alongside the rapid advancement of generative models~\citep{Jo2022ScorebasedGM,vignac2023digress,Chen2023ExploringTP,chen2025graphgenerativepretrainedtransformer}.
This field aims to address long-standing challenges stemming from the scarcity of high-quality datasets and the limitations in their secure and safe sharing processes. Synthetic data generation offers a promising research avenue, with the potential to enhance various aspects of \ml and \ai by enabling more robust, generalizable, and up-to-date analyses.

The expressivity of graph data entails substantial complexity, with each problem domain encoding unique soft and hard semantic constraints on what graphs are plausible and valid.
A recently developing line of research in \emph{synthetically generating graph-structured datasets} has begun to chip away at this complexity by considering constrained classes of graphs.
While many approaches have been proposed in the literature, most studies have focused on well-defined graphs with simple schema, typically involving nodes with only a single categorical or numerical attribute. These efforts have emphasized scalability, aiming to support very large graphs where the semantics of the graph are primarily determined by the existence of edges, rather than by detailed attributes.

% NOTE (@kjee) -- I don't the following paragraph much; Need another iteration to revise.
Modern graph applications in many domains have embraced complex heterogeneous graph schemas, including dozens of numeric, categorical, \emph{and textual} attributes for each node and edge.
For example, system provenance graphs --- widely used for cybersecurity applications --- include nodes that represent computer processes, and include semantically critical attributes such as the command that launched the process, the executable name, the PID, file location, the process owner, \etc
Further, provenance graphs include nodes for other system entities (files and network sockets) that have their own long lists of complex attributes, whose plausibility is intrinsically tied to the graph structure.

Existing synthetic graph generation research that focuses on simple structure and basic categorical and numeric attribute sets fails to adequately serve the needs of these emerging applications that deeply rely on the expressivity of complex graph schemas.
%However, widely adopted graph-structured datasets are failing to reflect such trend. They have rather focused on very large graphs with relatively simpler schema.

In this research, we propose \pname, a novel framework for generating synthetic graphs with a specific focus on supporting datasets with highly complex heterogeneous schemas. The central insight of \pname is that the graph generation problem can be cast as a sequence generation problem, for which there are ample techniques available in the literature. \pname is constructed by wrapping a sequence generation model in a graph-to-sequence encoder and sequence-to-graph decoder. For the inner sequence generation task, \pname leverages recent advances in transformer-based \llms.
Specifically, \pname's graph-to-sequence encoder/decoder has three design goals: \nm{1} lossless preservation and recovery of both the graph's structure and arbitrary serializable node and edge attributes; \nm{2} compact encoding of large graphs to efficiently utilize the sequence generation model's context window; and \nm{3} learnable single-pass generation of serialized graphs to maximize the effectiveness of the sequence generation model.

Leveraging the graph-to-sequence encoder, \pname first translates training graphs into compact sequence representations, capturing both graph structure and complex node and edge attributes. Then, these serialized graphs are used to fine-tune transformer-based \llms. Finally, \pname generates synthetic graphs by using the fine-tuned \llm to generate serialized graph sequences which are expanded into full graphs by the sequence-to-graph decoder.

\comment{%
    Graph schema can be more and more complicate as we attempt to capture complex and realistic phenomena using the graph format. In such cases, the graph schema would get more and more complicated. The baseline graph schema would be projected as heterogeneous graph format by having different node and edge types and graph nodes and edges are to be labelled with different types of attributes (\eg{} numerical or textual types).
    Widely adopted public datasets, while well representing important real-world problems with very large graph datasets, fail to reflect the complicated apsects of graph schema, which attempts to represent real-world problems in an accurate and detailed manners.
}

% TODO (@kjee): One line (or two) summary of prior work on synthetic graph generation research.

\comment{% TODO(@kjee) - we plan to have a section/subsection to introduce/describe system provenance graphs; Reserve the following for it.
    System provenance~\citep{inam2023sp} has gained prominence in recent years as a powerful tool to counter advanced cyberattack campaigns. The seminal work of King \etal{~\citep{king:2003sosp}} established system provenance as a vital field for comprehensive system-wide surveillance and protection. In system provenance analysis, causal graphs are extracted from surveillance logs by connecting resources through syscall-level access events. As outlined in \autoref{tab:schema}, in provenance graphs, nodes represent system entities (\eg processes, files, and network sockets), while edges denote the interactions between these entities, including \code{CREATE} a process, \code{READ} or \code{WRITE} a file, or \code{SEND} or \code{RECEIVE} data to or from a network socket. These nodes have security-relevant attributes such as filenames, executable names, and IP addresses. Provenance graphs are consumed by \ml models to perform security critical tasks, such as intrusion detection~\citep{rehman2024flash} and program classification~\citep{mukherjee2023sec}.
}

\comment{%
    \begin{figure}[h!]
    \centering
        \resizebox{0.9\textwidth}{!}{%
        \begin{subfigure}{0.40\textwidth}
            \centering
            \resizebox{\textwidth}{!}{%
            \includegraphics{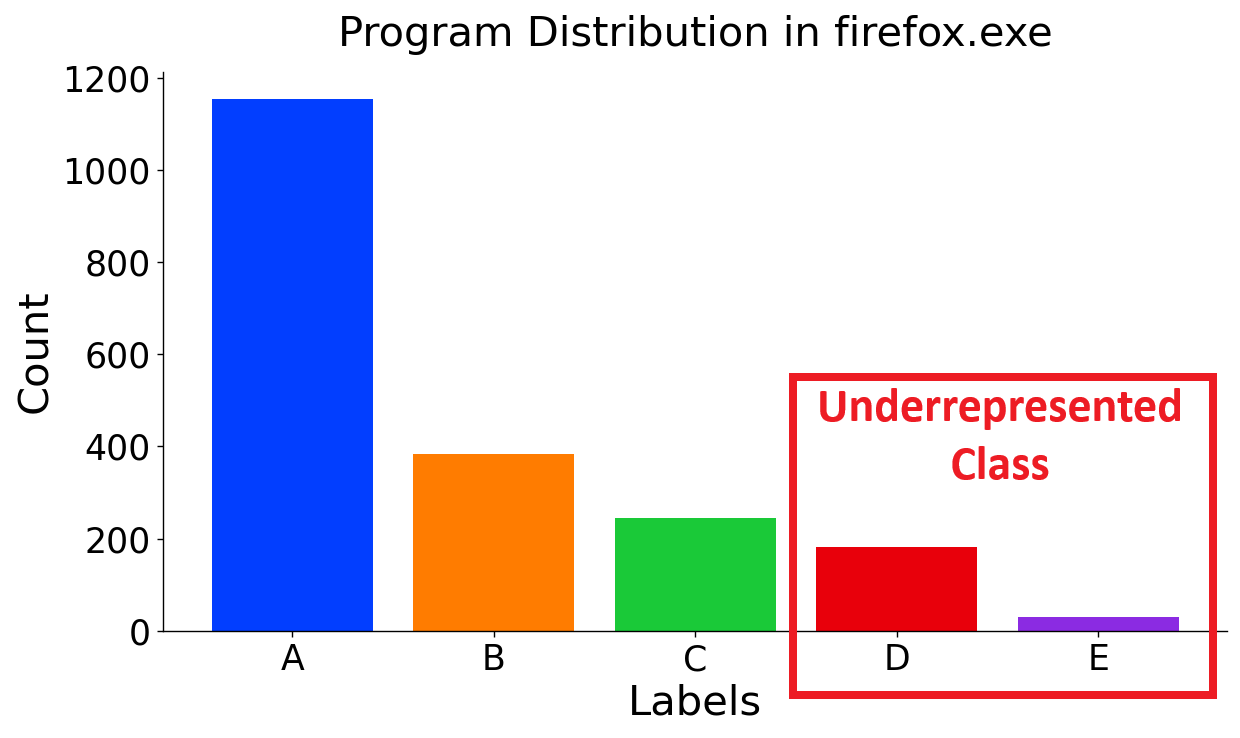}
            }
            \caption{\code{firefox.exe}}
            \label{fig:imbalance-firefox}
        \end{subfigure}
        \hfill
        \begin{subfigure}{0.40\textwidth}
            \centering
            \resizebox{\textwidth}{!}{%
            \includegraphics{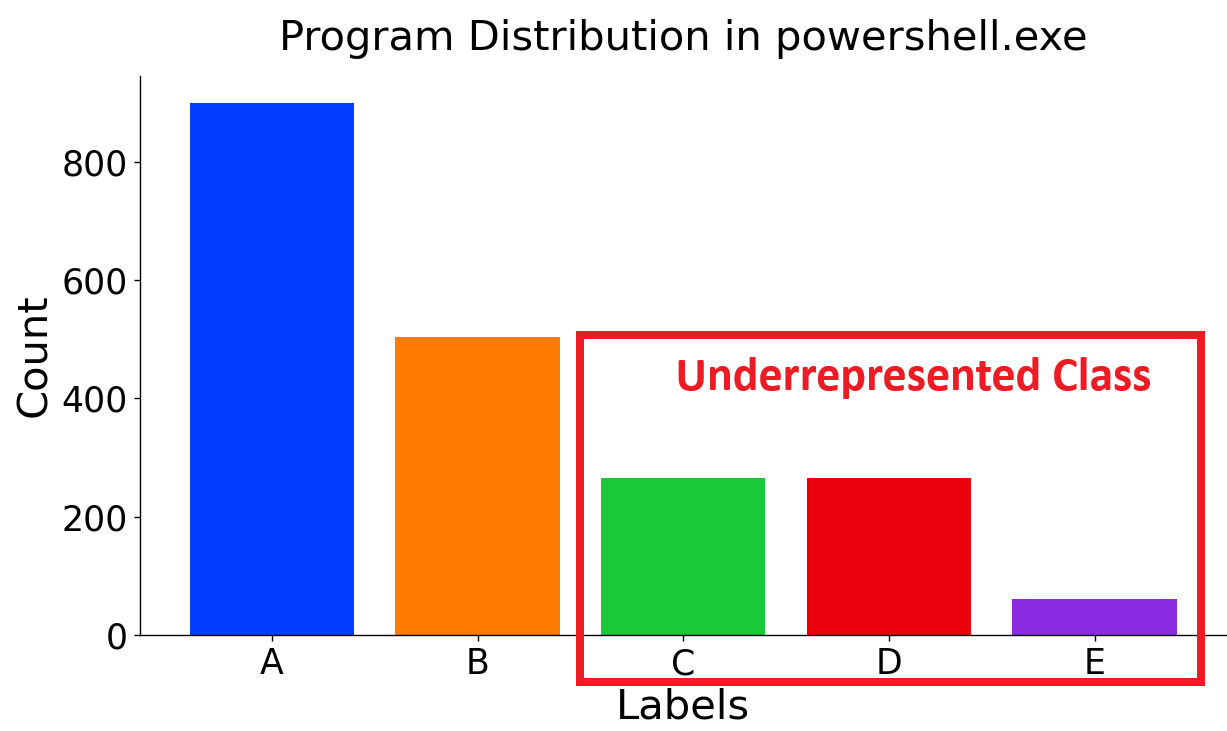}
            }
            \caption{\code{powershell.exe}}
            \label{fig:imbalance-powershell}
        \end{subfigure}
        }
        \caption{Provenance dataset of \code{svchost.exe} and \code{powershell.exe} showcase that certain programs are underrepresented in the dataset causing dataset imbalance. The underrepresented program being labelled D and E for \code{svchost.exe}, and C, D, and E for \code{powershell.exe}. For detailed descriptions of these labels, refer to \autoref{sec:label}.}
        \label{fig:imbalance}
    \end{figure}
}

\comment{%
    However, the effectiveness of \ml models is hampered by the inherent data imbalance in system provenance datasets, as shown in \autoref{fig:imbalance}. These datasets often lack sufficient representation of program behaviors, particularly when it comes to less common or underrepresented runtime configurations such as \code{WaaSMedicSvc} (label D) and \code{DoSvc} (label E) of \code{svchost.exe} and \code{Write-Host} (label C), \code{\$Infs = Get-Item} (label D), and \code{Get-AppxPackage} (label E) of \code{powershell.exe}. Such imbalance introduces systematic bias in the training process, leading to poor generalizability of security models. A balanced system provenance dataset is crucial for ensuring reliable intrusion detection. An ideal dataset should include equal representation of all relevant programs, as well as their various runtime configurations, such as command-line arguments. The imbalance found in real-world datasets means that \ml models trained on these datasets may fail to generalize to unseen environments~\citep{zhou2023improving}. This could reduce their effectiveness in practical deployments~\citep{alsaheel2021atlas} as models may incorrectly predict the behavior of underrepresented programs, leading to false positives in \ml -based intrusion detection systems.
}

\comment{%
    In this work, we introduce \pname, a novel graph synthesis framework designed to address the under-representation of program behaviors in system provenance datasets by synthesizing realistic provenance graphs. These synthetic graphs maintain both structural and attribute-wise similarity to the original data. To the best of our knowledge, this is the first approach within the provenance domain to simultaneously generate both graph structure and conditionally accurate node attributes. Outside the provenance domain, previous works such as \citet{Jo2022ScorebasedGM} have focused on generating graph structures with categorical attributes, but have not tackled the challenge of learning textual node attributes conditioned on graph structure --- an inherently complex task, as node attributes are influenced by both local neighborhood and distant graph nodes.
}

\comment{%
    Beyond just generating a uniform and fixed set of text attributes for all nodes, \pname offers the flexibility to associate nodes with arbitrary sets of text attributes, each with distinct semantics that can be incorporated and modified with no architecture changes. The ability to synthesize multiple text attributes for each node with distinct semantics has applications across many domains where graphs contain semantically rich attributes. While our work only considers static attribute sets for each node type, \pname's architecture naturally allows for dynamically determined node attribute sets using flexible attribute indicators.
}

\comment{%
    By jointly generating the graph structure and node attributes, \pname is able to achieve better structural and attribute-wise fidelity with the original training distribution compared to prior works, and is able to better support downstream model performance for both classification and intrusion detection.
}
\comment{%
    % By generating graphs that fill the gaps in program behavior representation, \pname helps to mitigate the data imbalance problem, thereby enabling more accurate and generalizable \ml models. Our evaluation demonstrates that \ml models trained on \pname-augmented balanced datasets achieve superior performance compared to models trained on imbalanced data. Specifically, models trained with augmented data show XX\% improvement in program classification and XX\% improvement in anomaly detection tasks, highlighting the utility of our augmentation framework in enhancing the robustness and accuracy of \ml models.
}

In summary, this work makes the following contributions:
\begin{itemize}[topsep=0pt,itemsep=-1ex,partopsep=1ex,parsep=1ex, leftmargin=*]
%MK-05-14-25: Now we claim we are general synthetic graph generator.  If so, we may rewrite the abstract and intro accordingly.
    \item \textbf{A unified framework for synthesizing complex heterogeneous graphs.} \pname introduces a generic framework to generate synthetic graph datasets with rich heterogeneity, including diverse node and edge types with high-dimensional attributes.

    \item \textbf{Joint modeling of structure and semantics.} \pname captures intricate dependencies between graph topology and attribute information by interleaving structural relationships with node and edge attributes during both encoding and generation phases.

    \item \textbf{A versatile graph-to-sequence encoder.} At the heart of \pname is a novel graph encoder that faithfully preserves structural and semantic information in a compact sequence format. This design enables seamless integration with transformer-based large language models and generalizes to a wide range of downstream tasks beyond synthetic graph generation.
\end{itemize}

To demonstrate the effectiveness and the general applicability of \pname, we evaluate it with datasets from two distinct domains. Firstly, we apply \pname approach to system provenance graphs from the cybersecurity domain, where the lack of up-to-date, high-quality public datasets has posed significant research limitations. Secondly, we evaluate \pname on knowledge graphs derived from the IntelliGraph Benchmark Dataset \citep{thanapalasingam2023intelligraphs}, which feature highly heterogeneous schemas and free-form attribute annotations.

% TODO (@kjee): summarize and highlight the key evaluation results here before we wrap up the intro.
To benefit the research community, we plan to publicly release our synthetic graph generation framework, as well as the datasets generated in this study after applying privacy-enhancing techniques to ensure data confidentiality and privacy.

%! root=../main.tex
\section{Background}\label{sec:back}

\subsection{Synthetic Data Generation.}
Synthetic data generation involves creating artificial data that closely resembles real data in its statistical properties and complexity without collecting additional real data. These approaches aim to address challenges presented by imbalanced datasets by supplementing underrepresented classes with synthetic examples. Recent advancements in data analysis techniques and data processing have led to numerous research proposals for data synthesis across various domains~\citep{8953317,9209598,10.1145/3424155,feng2021survey,10.1038/s41467-022-35295-1,zhao2020data,zhao2022arxiv}.
Existing synthetic graph generation methods primarily focus on single categorical attributes during graph structure generation (e.g., molecular types for nodes and bond types for edges). However, capturing more complex attributes requires additional mechanisms to model their interdependencies. Most prior work addresses this by generating vector representations of node features~\citep{li2024graphmakerdiffusionmodelsgenerate,li2024efficientdynamicattributedgraph}. We argue and empirically demonstrate that node attributes and graph structure should be generated jointly, as they are often statistically interdependent. To this end, \pname formulates both as a unified optimization problem. Departing from traditional approaches in graph generation, we leverage recent advances in generative transformers for sequences, adapting them to the graph domain through carefully designed serialization and deserialization strategies. By generating graphs as constrained sequences, \pname enables unprecedented flexibility in modeling complex, interdependent node and edge attributes alongside the graph structure.

\subsection{System Provenance Graphs}
System provenance tracks fine-grained system data (\code{syscall} events), from large enterprise and industrial systems. It traces bidirectional information flow and control dependencies starting from a \poi event, enabling forensic analysis and advanced security defenses~\citep{king:2003sosp,liu2018ndss,hassan2019ndss,wang2020ndss}. By examining system-call logs~\citep{auditd,etw,dtrace2005lisa}, system provenance graphs capture relationships (\ie \code{READ}, \code{WRITE}, \code{CREATE} and \code{EXECUTE}) among major resources (\ie processes, files and network sockets). Nodes, representing system resources, are annotated with attributes such as executable names, filenames, and IP addresses, making these graphs invaluable tools for forensic analysis to discover points of entry, track infection propagation, and assess the scale of damage. Formally, a provenance graph is a connected set of timestamped edges $e = (u, v, r)$, where $u, v \in \{processes \cup files \cup sockets\}$ and $v$ is causally dependent on $u$ (\eg a file $u$ is written by a process $v$), and $r$ is the relationship between the nodes.

%MK-05-14-25: I wondwer the following paragraph makes more sense after the system provenance is introduced.
Although synthetic graph generation research has attracted significant attention across various research domains, security datasets, particularly system provenance graphs, have seen limited progress. Synthetic data is valuable in the system provenance domain because real \code{syscall} traces are costly to collect and difficult to share due to privacy concerns, but are required in large quantities by \ml models. \citet{creech2013generation} and \citet{haider2016fi} proposed initial prototype implementations for synthesizing attack behaviors by generating sequences of system calls, but these implementations lacked consideration of causal dependencies among system resources (\eg processes, files, and network sockets). The operating system (OS) used in \citet{creech2013generation} is decade old Ubuntu 11.04 which has reached end of life support and it contains no Windows traces. Furthermore, \citet{creech2013generation} rejected traces over 3 kB thus impacting the completeness of the traces collected. By directly generating provenance graphs, \pname synthesizes data in an OS-agnostic way with structural consideration of causal dependencies, and can generate large graphs by combining related subgraphs. Notably, edges in provenance graphs are naturally ordered by the time at which \code{syscall} events occur, which synergizes with \pname's sequential generation approach. While synthetic data generation has been explored in other security research domains~\citep{10.1145/3424155}, we are unaware of any previous research focusing on synthetic data generation approaches specifically for system provenance graph datasets.

\subsection{Provenance-based \ml Research}
Initially proposed to automate forensic investigations, system provenance has become a vital foundation for \ml-based security detectors~\citep{wang2020ndss,han2020ndss,han2021sigl,jia2023magic, cheng2023kairos, mukherjee2024proviot, rehman2024flash}, propelled by rapid advances in data analysis techniques. Security researchers have expanded the scope of system provenance studies to construct \ml models that counter attacks where adversaries craft unique, unexposed attack vectors~\citep{mukherjee2023sec} that can only be identified during runtime. While the full provenance graph even for one host machine can be intractably large, comprising millions of edges, analysis can be performed on tractable subgraphs extracted by tracing causal dependencies to and from a \poi event. Depending on the objective of the analysis task, various query conditions can determine effective \poi events, with the system provenance graph serving as the primary target for analysis.
%

% TODO (@kjee) --- this can be relocated near to eval section.
%MK-05-14-25: the next sentence does not make sense. Please move to the eval section.  I commented out for now.
%\subsection{Highly Complex Graph Datasets}\label{sec:graph_datasets}
%In this research, to validate \pname approach, we incorporate two different type of datasets.

%! root=../main.tex
\section{Graph Synthesis Methodology}\label{sec:methodology}

\begin{figure}[!tb]
    \centering
    \includegraphics[width=.8\linewidth]{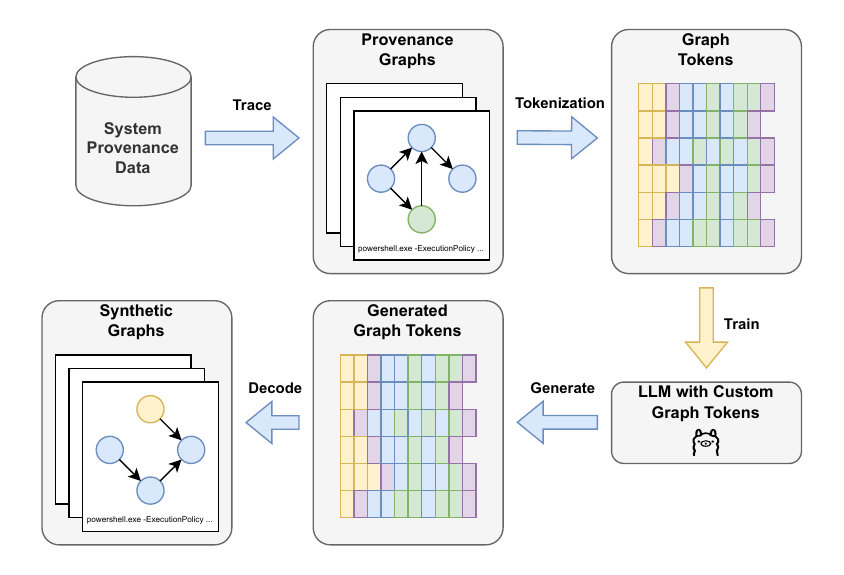}
    \caption{\pname's synthetic graph generation pipeline.}
    \label{fig:overview}
\end{figure}

%\subsection{\pname Workflow}\label{sec:workflow}
\pname focuses on two key aspects of graph generation: the generation of the graph structure and the prediction of node attributes within the generated graph.
Previous research has primarily faced graph structure and attribute generation as separate problems, without attempting structural and attribute generation into a single generative process. Our work is the first to integrate both approaches targeting graphs with highly complex node and edge attributes in the cybersecurity domain. System provenance graphs provide a uniquely challenging scenario with heterogenous graph structure and varied node attributes such as file paths, process command line arguments, and IP addresses. These attributes are not only complex but also interdependent with the graph structure, making an ideal use case for a method that generates both jointly.

\pname models graph generation as a sequence generation task, where each graph is represented as a sequence of tokens. This enables the use of an LLM as the backbone of our architecture, allowing us to leverage the power of pretrained models for increased performance and faster training time than training a model from scratch. We use LLama3-3.2-3B as the base model for \pname. Additionally, we propose an efficient graph representation which addresses the challenges of generating long sequences of tokens while ensuring the validity of the generated graph structure.
Utilizing the LLM backbone allows us to use prompts to condition the graph generation process, enabling the generation of graphs that are relevant to specific contexts, such as different command line arguments being passed to a process. This is particularly useful in cybersecurity applications where the context can significantly influence program behavior, thereby changing the structure and attributes of the generated graph.

The overall workflow of \pname is shown in \autoref{fig:overview}.

\subsection{Graph Representation}

Efficiently representing a graph is crucial for the generation process. Previous works such as \cite{Chen2023ExploringTP} and \cite{wang2023can} consider representing graphs in plain-text formats. This encoding is inefficient due to the use of many tokens to represent information that is not critical to the graph structure, while also presenting challenges for reconstructing graphs from model outputs. As the graph grows large, the chance of generating syntactically invalid tokens increases, leading to difficulties in decoding the generated tokens back into a valid graph structure. This is particularly problematic for heterogeneous graphs, where different node and edge types may have distinct attributes and relationships, making it difficult to ensure that the generated tokens conform to the expected structure.

To reduce the overhead of text representation, we designed a list of special tokens for efficient graph representation, described in \autoref{tab:tokens}. The graph is wrapped inside a pair of tokens \textless bog\textgreater, \textless eog\textgreater{} indicating the start and end of the graph. We inject $N$ node tokens to represent up to $N$ nodes. Edges are represented as pairs of nodes. Since our dataset is heterogeneous, we also need to represent the node types and edge types. We use \textless ntype$_i$\textgreater{} and \textless etype$_i$\textgreater{} to represent the node type and edge type respectively. The node attributes are represented as a sequence of tokens, where each attribute is wrapped inside a pair of tokens \textless bof\textgreater{} and \textless eof\textgreater{}. This allows us to easily identify the start and end of each attribute. Comparing to directly representing the graph in a plain text format, such as the JSON node-link data format in python's NetworkX library, our tokenization schema only uses $35\%$ and $22\%$ tokens on provenance data and IntelliGraph data respectively.

\begin{wraptable}{r}{0.45\textwidth}
    \centering
    \caption{Tokenization schema for graph representation.}
    \label{tab:tokens}
    \resizebox{0.4\columnwidth}{!}{
    \begin{tabular}{@{}ll@{}}
    \toprule
    \textbf{Token} & \textbf{Description} \\ \midrule
    \textless bog\textgreater        & Beginning of graph   \\
    \textless eog\textgreater        & End of graph         \\
    \textless bon\textgreater        & Beginning of node    \\
    \textless eon\textgreater        & End of node          \\
    \textless boe\textgreater        & Beginning of edge    \\
    \textless node$_i$\textgreater     & Node $i$             \\
    \textless ntype$_j$\textgreater    & Node type $j$        \\
    \textless etype$_k$\textgreater    & Edge type $k$        \\
    \textless bof\textgreater        & Beginning of node or edge attribute  \\
    \textless eof\textgreater        & End of node or edge attribute    \\
    \bottomrule
    \end{tabular}%
    }
\end{wraptable}

To encode each distinct graph into a unique sequence of tokens, we assume that there exists a well-defined order to visit every edge in the graph. In our provenance dataset, each edge represents an action by a program with an associated timestamp, which can be used to naturally determine the order in which edges are encoded.
For general graphs, we use a topological sort to determine the order of the edges.
A strict ordering is important because it allows us to generate the graph in a way that respects the dependencies between nodes and edges, while reducing the ambiguity in the generated sequence of tokens.
When encoding the graph, we append each edge in the graph following this pre-defined order.
Whenever a new edge is added to the sequence that connects to any node not yet seen in the graph, we append the node attribute tokens immediately after.

The detailed process is described in \autoref{alg:tokenization} in the Appendix.

\autoref{fig:graph_tokens} shows an example of how a graph is represented as a sequence of tokens.

%MK-05-14-25: Future update ?: I think this figure may used similar node and features such as process, file, proc_create etc on the tokenization example so left graph matches exactly with the righ tokenization example for clarity.
\begin{figure}[h]
    \centering
    \includegraphics[width=.9\linewidth]{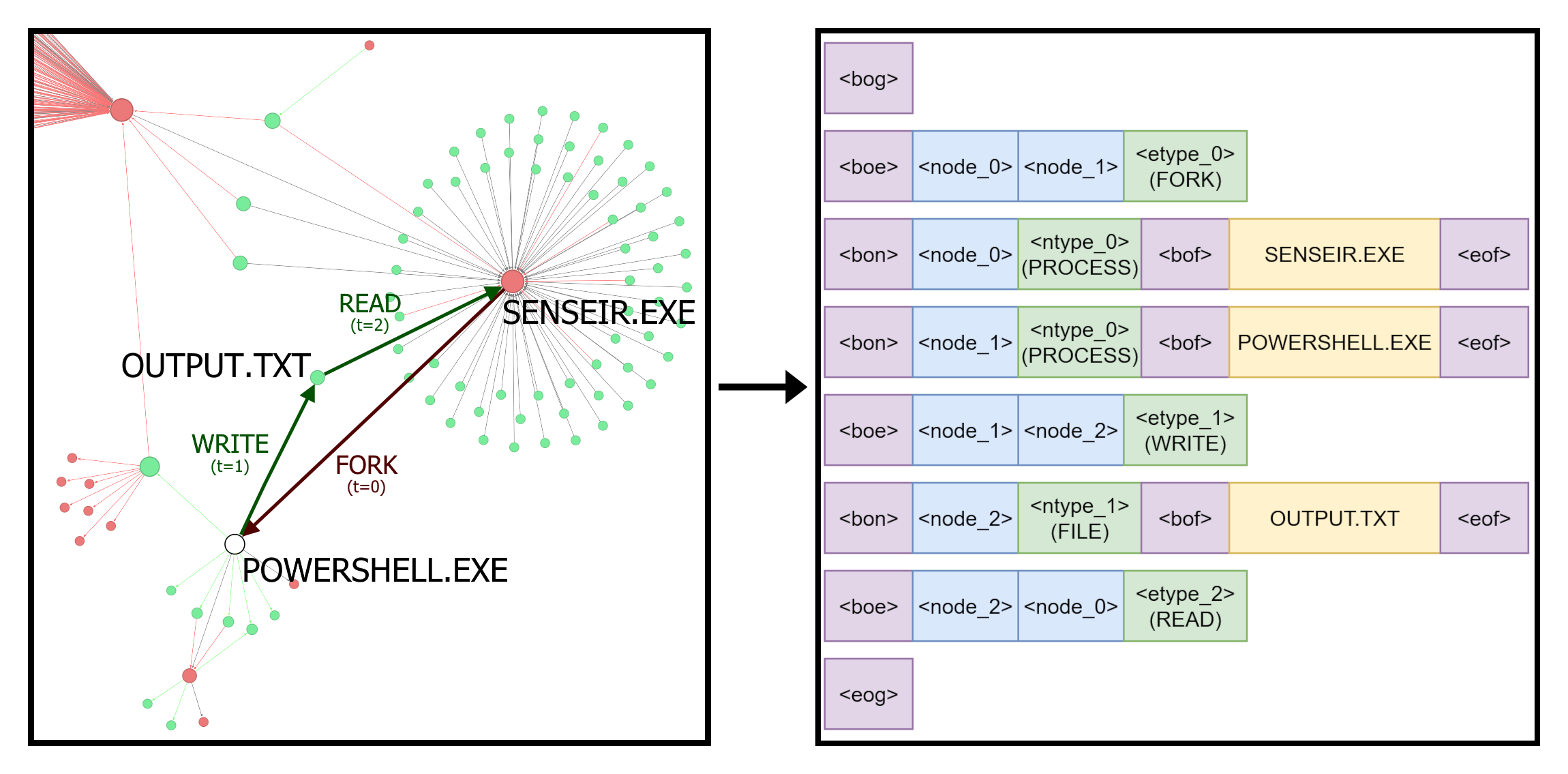}
    \caption{Example of graph tokenization. The left side of the figure highlights a subgraph of a real provenance graph containing a \code{powershell.exe} process. The right side of the figure shows the corresponding tokenized sequence using our proposed schema and time-based edge ordering. File and process names are shortened.}
    \label{fig:graph_tokens}
\end{figure}

\subsection{Decoding Tokens to Graph}

When sampling long outputs from LLM, the model may generate invalid tokens or token sequences that do not follow the expected graph structure. To address this issue, we employ two methods to ensure the validity of the generated tokens.

Firstly, we utilize a token filtering mechanism that checks for invalid tokens or sequences during the generation process. We enforce a set of simple rules during sampling --- for example, only node tokens \textless node$_i$\textgreater{} are allowed to be followed by begin-of-node tokens \textless bon\textgreater, so we only sample from node tokens. This filtering step helps to eliminate any tokens that do not conform to the expected graph structure or violate the constraints of the graph representation.

Secondly, we designed anchor tokens \textless{}boe\textgreater{}, \textless{}bon\textgreater{} and \textless{}bof\textgreater{} to help with the parsing process. We implemented the parser with a state machine. When an invalid token is encountered, the parser attempts to recover by skipping the invalid token and continuing to parse the remaining tokens. When an anchor token is encountered it can directly trigger a state transition in the parser, allowing the parser to recover to a valid state. This approach allows us to salvage as much of the generated graph as possible, even in the presence of errors or unexpected tokens.

Combining the filtering and decoding steps, we can effectively handle the challenges associated with generating long sequences of tokens while maintaining the validity of the graph structure.
% \txw{We compare the performance of our tokenization and decoding approach with the plain text/json representation in XXX and show that our approach significantly reduces the number of invalid graphs and improves the overall quality of the generated graphs.}

\subsection{Efficient Training for LLM based Graph Generation}
Utilizing large language model architecture for graph generation is a promising approach. However, it also presents challenges, particularly in terms of training efficiency. The training process can be computationally expensive and time-consuming, especially when dealing with large graphs. In this work, we adopt LLaMA3-3.2B as our backbone, a lightweight and efficient LLM architecture with a reasonable trade-off between performance and resource consumption.
To accommodate our graph-specific tokenization, we initialize with the pretrained model weights and expand the embedding layer using mean-resizing to incorporate new special tokens. This strategy preserves the pretrained knowledge while enabling the model to learn representations for the added tokens.
To further improve training efficiency, we employ mixed-precision training and LoRA-based fine-tuning, allowing us to harness the capabilities of large language models while significantly reducing computational overhead.

\section{Evaluation}

\heading{Evaluation Protocol and Datasets.}
We train \pname to synthesize provenance graphs for \code{firefox.exe} and \code{powershell.exe} on Windows, and use the resulting graphs to augment under-represented sub-program classes in our testing corpus. These two programs are frequent targets of process-impersonation attacks by advanced cyber-threat actors~\citep{survivialism2021sp}; moreover, their behavior is largely driven by command-line arguments, which yields clear class labels for evaluation. Our training data consists of system provenance graphs generated by forward-tracking from 50\,492 instances of \code{powershell.exe} and 7\,901 instances of \code{firefox.exe} representing real user and system activity collected from desktop systems actively in use by our lab.

We compare our model against the \gdss model~\citep{Jo2022ScorebasedGM}, which is a state-of-the-art diffusion based graph generation model based on score-based generative modeling. Since the implementation from the original paper does not scale well on very large graphs, we used a third-party graph-transformer based implementation, and we adjusted the hyperparameters to handle our dataset. The details are provided in the appendix.

For the structural fidelity studies (\autoref{sec:eval-structure-fidelity}, \autoref{sec:eval-structure-fidelity-intelligraph}, and \autoref{sec:eval-embedding-fidelity}), we sample 1\,000 graphs for each of \code{firefox.exe} and \code{powershell.exe} from \pname and from the score-based generator of \citet{Jo2022ScorebasedGM}, both trained on the same data. We then quantify the distributional distance between generated and training graphs with established graph-similarity metrics. 

%To capture provenance-specific node semantics more faithfully, we introduce a domain-aware node-attribute similarity measure, \bleup, detailed in \autoref{sec:appx-bleu-plus}. Because the model of \citet{Jo2022ScorebasedGM} does not natively emit textual node attributes, we equip its outputs with randomly sampled attributes drawn from the training set, thereby providing a competitive baseline.

For downstream tasks (\autoref{sec:eval-downstream}), we refer to respected works from the cybersecurity domain to guide our supervised program classification task \citep{survivialism2021sp}. Two identical \gnn models are trained, one with a training set of real graphs and one with an identical number of synthetic graphs generated with \pname. \gnn models are trained separately to classify different command line arguments for both \code{firefox.exe} and \code{powershell.exe}. Both models are evaluated on real and synthetic test data.

In addition, to evaluate the general performance of \pname, we also evaluated it on the \textit{IntelliGraph} benchmark~\citep{thanapalasingam2023intelligraphs}, which is a collection of synthetic and real-world graph datasets. The benchmark includes three synthetic datasets and two real-world datasets, and we compare our model against the baseline models provided in the benchmark.

\newcommand{\SvchostOldGDSSDegree}{0.038}
\newcommand{\SvchostOldGDSSCluster}{0.022}
\newcommand{\SvchostOldGDSSBetCen}{0.008}
\newcommand{\SvchostOldGDSSCloseCen}{0.228}
\newcommand{\SvchostOldGDSSKatzCen}{0.209}
\newcommand{\SvchostOldGDSSSpectral}{0.115}

\newcommand{\SvchostOldProvcreatorDegree}{\textbf{0.030} (\emph{-0.008})}
\newcommand{\SvchostOldProvcreatorCluster}{\textbf{0.002} (\emph{-0.020})}
\newcommand{\SvchostOldProvcreatorBetCen}{\textbf{0.007} (\emph{-0.001})}
\newcommand{\SvchostOldProvcreatorCloseCen}{\textbf{0.175} (\emph{-0.053})}
\newcommand{\SvchostOldProvcreatorKatzCen}{\textbf{0.068} (\emph{-0.141})}
\newcommand{\SvchostOldProvcreatorSpectral}{\textbf{0.060} (\emph{-0.055})}

\newcommand{\FirefoxGDSSDegree}{0.955}
\newcommand{\FirefoxGDSSCluster}{0.130}
\newcommand{\FirefoxGDSSBetCen}{0.535}
\newcommand{\FirefoxGDSSCloseCen}{0.245}
\newcommand{\FirefoxGDSSKatzCen}{0.512}
\newcommand{\FirefoxGDSSSpectral}{0.410}

\newcommand{\FirefoxProvcreatorDegree}{\textbf{0.277}}
\newcommand{\FirefoxProvcreatorCluster}{\textbf{0.095}}
\newcommand{\FirefoxProvcreatorBetCen}{\textbf{0.138}}
\newcommand{\FirefoxProvcreatorCloseCen}{\textbf{0.150}}
\newcommand{\FirefoxProvcreatorKatzCen}{\textbf{0.313}}
\newcommand{\FirefoxProvcreatorSpectral}{\textbf{0.247}}

\newcommand{\PowershellOldGDSSDegree}{0.013}
\newcommand{\PowershellOldGDSSCluster}{0.267}
\newcommand{\PowershellOldGDSSBetCen}{0.010}
\newcommand{\PowershellOldGDSSCloseCen}{0.164}
\newcommand{\PowershellOldGDSSKatzCen}{0.064}
\newcommand{\PowershellOldGDSSSpectral}{0.066}

\newcommand{\PowershellOldProvcreatorDegree}{\textbf{0.003} (\emph{-0.010})}
\newcommand{\PowershellOldProvcreatorCluster}{\textbf{0.048} (\emph{-0.219})}
\newcommand{\PowershellOldProvcreatorBetCen}{\textbf{0.007} (\emph{-0.003})}
\newcommand{\PowershellOldProvcreatorCloseCen}{\textbf{0.094} (\emph{-0.070})}
\newcommand{\PowershellOldProvcreatorKatzCen}{\textbf{0.044} (\emph{-0.020})}
\newcommand{\PowershellOldProvcreatorSpectral}{\textbf{0.021} (\emph{-0.045})}

\newcommand{\PowershellGDSSDegree}{1.148}
\newcommand{\PowershellGDSSCluster}{0.487}
\newcommand{\PowershellGDSSBetCen}{0.295}
\newcommand{\PowershellGDSSCloseCen}{1.213}
\newcommand{\PowershellGDSSKatzCen}{1.276}
\newcommand{\PowershellGDSSSpectral}{1.022}

\newcommand{\PowershellProvcreatorDegree}{\textbf{1.109}}
\newcommand{\PowershellProvcreatorCluster}{\textbf{0.158}}
\newcommand{\PowershellProvcreatorBetCen}{\textbf{0.147}}
\newcommand{\PowershellProvcreatorCloseCen}{\textbf{0.997}}
\newcommand{\PowershellProvcreatorKatzCen}{\textbf{1.251}}
\newcommand{\PowershellProvcreatorSpectral}{\textbf{0.915}}

\begin{table}[h]
    \centering
    % \txw{Not a trivial task, need to update all the baselines models to work with provenance data}
    % \kunal{@tianhao: add citations for transE, distmult, complex}
    \caption{Maximum mean discrepancy (MMD) distances of different graph metrics between \inhouse and synthetic datasets generated with \gdss~\citep{Jo2022ScorebasedGM} and \pname where lower number is better.
    %\kunal{9-15 Discussion with @Tianhao: We cannot distinguish between metrics of different node type, because the metric assumes that the nodes of a particular node types are connected, which is not always true. While we can get the average value for node types but they are meaningless.}
    }
    \label{tab:structure}
    \resizebox{0.85\columnwidth}{!}{
    \begin{tabular}{@{}lccccccc@{}}
    \toprule
      & 
    Degree $\downarrow$ &
    Clustering $\downarrow$ & 
    Bet. Cen. $\downarrow$ &
    Cls. Cen. $\downarrow$ &
    Katz Cen. $\downarrow$ &
    Spectral $\downarrow$ \\ 
    \cmidrule(lr){2-7} 
    & 
    \multicolumn{6}{c}{\code{firefox.exe}} \\
    \midrule
    % \inhouse          & xx.xx ($\pm$ xx.xx) & xx.xx ($\pm$ xx.xx) & xx.xx ($\pm$ xx.xx) & xx.xx ($\pm$ xx.xx) & xx.xx ($\pm$ xx.xx) & xx.xx ($\pm$ xx.xx) \\
    \gdss          & \FirefoxGDSSDegree & \FirefoxGDSSCluster & \FirefoxGDSSBetCen & \FirefoxGDSSCloseCen & \FirefoxGDSSKatzCen & \FirefoxGDSSSpectral \\
    % \random          & xx.xx & xx.xx & xx.xx & xx.xx & xx.xx & xx.xx \\
    % \transe          & xx.xx & xx.xx & xx.xx & xx.xx & xx.xx & xx.xx \\
    % \distmult          & xx.xx & xx.xx & xx.xx & xx.xx & xx.xx & xx.xx \\
    % \complex          & xx.xx & xx.xx & xx.xx & xx.xx & xx.xx & xx.xx \\
    % \midrule
    \textbf{\pname}     & \FirefoxProvcreatorDegree & \FirefoxProvcreatorCluster & \FirefoxProvcreatorBetCen & \FirefoxProvcreatorCloseCen & \FirefoxProvcreatorKatzCen & \FirefoxProvcreatorSpectral \\
    \midrule
     &
    \multicolumn{6}{c}{\code{powershell.exe}} \\
    \midrule
    % \inhouse          & xx.xx ($\pm$ xx.xx) & xx.xx ($\pm$ xx.xx) & xx.xx ($\pm$ xx.xx) & xx.xx ($\pm$ xx.xx) & xx.xx ($\pm$ xx.xx) & xx.xx ($\pm$ xx.xx) \\
    \gdss          & \PowershellGDSSDegree & \PowershellGDSSCluster & \PowershellGDSSBetCen & \PowershellGDSSCloseCen & \PowershellGDSSKatzCen & \PowershellGDSSSpectral \\
    % \random          & xx.xx & xx.xx & xx.xx & xx.xx & xx.xx & xx.xx \\
    % \transe          & xx.xx & xx.xx & xx.xx & xx.xx & xx.xx & xx.xx \\
    % \distmult          & xx.xx & xx.xx & xx.xx & xx.xx & xx.xx & xx.xx \\
    % \complex          & xx.xx & xx.xx & xx.xx & xx.xx & xx.xx & xx.xx \\
    % \midrule 
    \textbf{\pname}     & \PowershellProvcreatorDegree & \PowershellProvcreatorCluster & \PowershellProvcreatorBetCen & \PowershellProvcreatorCloseCen & \PowershellProvcreatorKatzCen & \PowershellProvcreatorSpectral \\
    \bottomrule
    \end{tabular}%
    }
\end{table}

\subsection{Structure Fidelity}\label{sec:eval-structure-fidelity}
\autoref{tab:structure} reports structural fidelity using multiple graph‑level statistics for the \inhouse and the corresponding synthetic dataset, mirroring the evaluation protocol of \citet{Jo2022ScorebasedGM}. We compute the \emph{Maximum Mean Discrepancy} (MMD) for different structural metrics between real graphs and those synthesized by \pname and the \gdss baseline, with both models trained independently on the \code{firefox.exe} and \code{powershell.exe} datasets. Across both programs, \pname achieves consistently lower MMD scores than \gdss, indicating superior structural fidelity. We attribute these gains to the learning ability of the transformer, which enhances disambiguation among graph categories and enhanced generation capability due to the auto-regressive nature of \pname.

\subsection{Attribute Fidelity}\label{sec:eval-attribute-fidelity}
\autoref{tab:attrib-validation} presents the results of the attribute fidelity evaluation. Our model generates both the graph structure and attributes simultaneously, making it challenging to directly compare against ground truth attribute values. To address this, we developed a set of custom rules to validate the generated attributes. Details of these rules are provided in \autoref{tab:attrib-validation-regex} in the appendix.

\begin{table}[h]
    \centering
    \caption{Valid rates for various attributes.}
    \label{tab:attrib-validation}
    \resizebox{0.85\columnwidth}{!}{
    \begin{tabular}{lcccc}
        \toprule
        Dataset & \% Valid Process Names $\uparrow$ & \% Valid File Names $\uparrow$ & \% Valid IP Address and Ports $\uparrow$ \\
        \midrule
        Firefox & 93.77 & 99.22 & 96.43 \\
        Powershell & 94.83 & 98.70 & 98.92 \\
        \bottomrule
    \end{tabular}
    }
\end{table}

\begin{table}[h]
    \centering
    \caption{Evaluation results using the IntelliGraph Benchmark Dataset (only the best performing baselines are shown).}
    \label{tab:structure-intelligraph}

    \resizebox{0.95\columnwidth}{!}{
    \begin{threeparttable}
    \begin{tabular}{@{}llcccc@{}}
    \toprule
    Dataset & 
    Model &
    \% Valid Graphs $\uparrow$& 
    \% Novel \& Vaild Graphs $\uparrow$ &
    \% Novel Graphs $\uparrow$ &
    \% Empty Graphs $\uparrow$ \\
    \midrule
    \multirow{2}{*}{syn-paths}
        & ComplEx (Sampling $P(E)$ and $P(S|E)$) & 0.71 & 0.71 & 14.27 & 85.73 \\
        & ComplEx (Sampling only $P(E)$) & 10.10 & 10.10 & 95.58 & 4.42 \\
        & \textbf{ProvCreator} & 93.14 & 93.14 & 100.0 & 0.0 \\
    \midrule
    \multirow{2}{*}{syn-tipr}
        & DistMult (Sampling $P(E)$ and $P(S|E)$) & 0.00 & 0.00 & 13.34 & 86.66 \\
        & ComplEx (Sampling only $P(E)$) & 0.00 & 0.00 & 99.64 & 0.36 \\
        & \textbf{ProvCreator} & 0.00 & 0.00 & 100.00 & 0.00 \\
    \midrule
    \multirow{2}{*}{syn-types}
        & TransE (Sampling $P(E)$ and $P(S|E)$) & 0.21 & 0.21 & 15.44 & 84.56 \\
        & DistMult (Sampling only $P(E)$) & 1.44 & 1.44 & 96.19 & 4.81 \\
        & \textbf{ProvCreator} & 86.16 & 86.16 & 100.00 & 0.00 \\
    \midrule
    \multirow{2}{*}{wd-movies}
        & ComplEx (Sampling $P(E)$ and $P(S|E)$) & 0.00 & 0.00 & 1.87 & 98.13 \\
        & ComplEx (Sampling only $P(E)$) & 0.41 & 0.41 & 93.04 & 6.96 \\
        & \textbf{ProvCreator} & 94.26 & 94.26 & 100.0 & 0.00 \\
    \midrule
    \multirow{2}{*}{wd-articles}
        & TransE (Sampling $P(E)$ and $P(S|E)$) & 0.00 & 0.00 & 4.58 & 95.42 \\
        & ComplEx (Sampling only $P(E)$) & 0.00 & 0.00 & 100.00 & 0.00 \\
        & \textbf{ProvCreator} & 0.00 & 0.00 & 100.00 & 0.00 \\
    \bottomrule
    \end{tabular}

    \begin{tablenotes}
        \small
        \item[*] Note: \pname is accomplishing a more difficult task while still outperforming the baselines. The baseline models does not need to generate the attributes as text, but instead sample the attributes as a set of numbers.
        \item[**] Due to very strict evaluation criteria on certain datasets, both our model and the baselines may not produce a fully valid graph, leading to zero scores in some cases. We reported the scores as-is, but the actual performance of our model may be better than indicated.
    \end{tablenotes}
    \end{threeparttable}
    }
\end{table}

%MK-05-14-25: the table 3 has some columns with zero 0.00. why ?? Also, Please high the winning entries.
\subsection{Evaluation on IntelliGraph Benchmark Dataset}
\label{sec:eval-structure-fidelity-intelligraph}
\autoref{tab:structure-intelligraph} shows the results when evaluating our model on the \textit{IntelliGraph} Datasets~\citep{thanapalasingam2023intelligraphs}. These datasets are made up of knowledge graphs where each node represents some entity and has a textual label, and each edge represents a relationship between nodes with a categorical relationship type. \citet{thanapalasingam2023intelligraphs} provide evaluation results for baseline graph generation models on each dataset, which we compare against \pname in \autoref{tab:structure-intelligraph}. These baseline methods generate node labels by sampling labels from the original dataset, whereas \pname generates node labels as textual node attributes. We also generate the edge type as a textual attribute, falling back to the first edge type in the dataset if the generated text is invalid. Our model outperforms the provided baseline models by a large margin, indicating that our model is capable of generating diverse and realistic graphs on general graph datasets.

% \input{tables/attribute.tex}
% \subsection{Attribute Fidelity}\label{sec:eval-attribute-fidelity}

\subsection{Embedding Fidelity}\label{sec:eval-embedding-fidelity}

\begin{figure}[H]
    \centering
    \includegraphics[width=.7\linewidth]{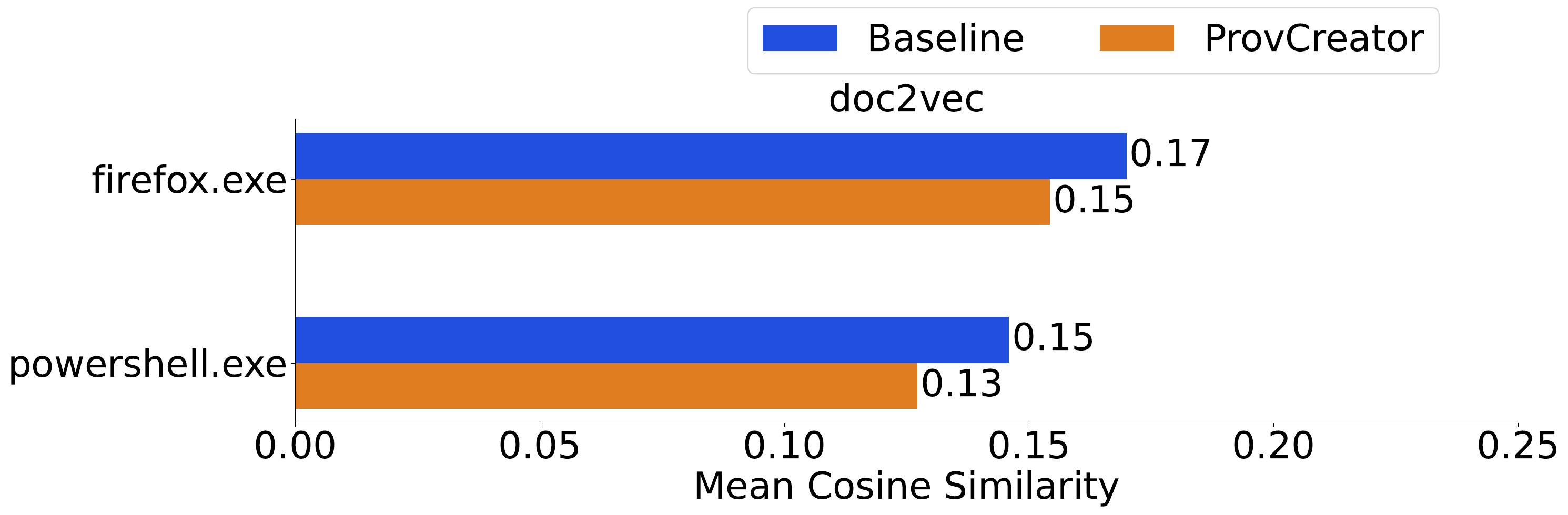}
    \caption{Embedding cosine similarity of real and synthetic datasets generated with \pname and a GDSS \citep{Jo2022ScorebasedGM} baseline, using doc2vec graph embeddings \citep{le2014distributed}.}
    \label{fig:embedding}
\end{figure}

We evaluate the overall similarity of the synthetic graphs to the original training set by embedding the graphs into vectors, then measuring the average cosine similarity of those graph embeddings to the embeddings of the training set. \autoref{fig:embedding} compares graphs generated by \pname with a baseline method. The baseline method uses real graph structures sampled from the training set, augmented with randomly sampled node attributes for parity. To embed the graph with text attributes into vectors, we sample $100$ random walk paths with maximum length $10$ on each graph, and log the nodes and edges visited along with the attributes to obtain a document representing the graph. Then we use doc2vec~\citep{Le2014} to obtain the embedding. \pname attains comparative similarity scores, demonstrating the ability to capture the relationship between the graph structure and node attributes.

\begin{figure}[H]
    \centering
    \includegraphics[width=0.7\linewidth]{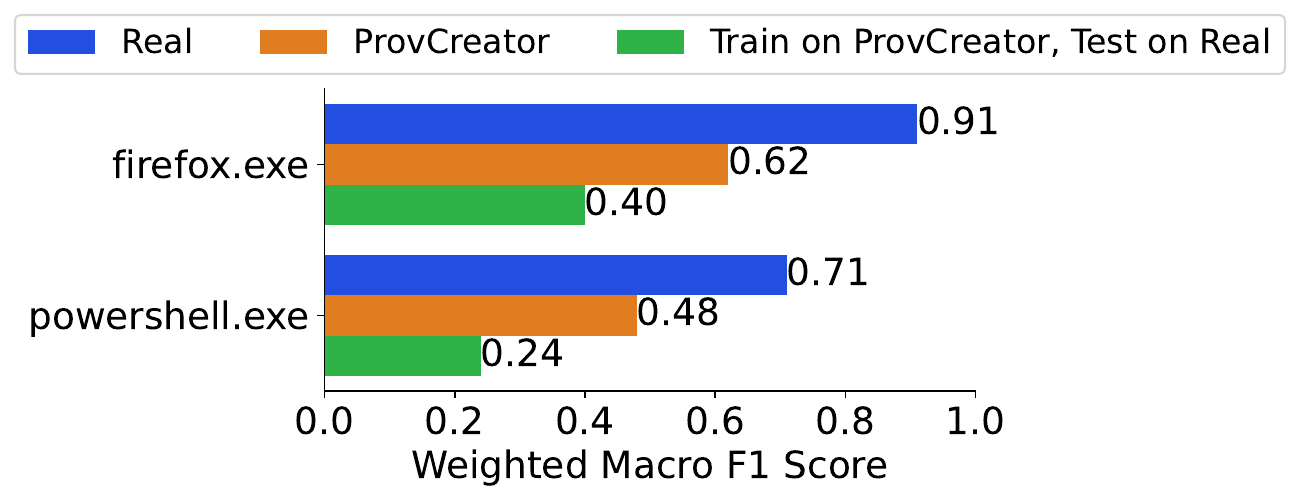}
    \caption{Weighted macro-F1 scores for GNN-based subprogram classification using different graph types.}
    % \kunal{@Tianhao: please confirm this is what we discussed? For Program classification, we will train on the generated (synthetic) graphs, and test on real graphs. Report the performance degradation versus models trained and tested on real graphs.}
    \label{fig:classification}
\end{figure}

\subsection{Downstream Application}\label{sec:eval-downstream}
To demonstrate the practical utility of \pname, we consider security-relevant downstream tasks of program classification (\autoref{fig:classification} show classification results). For both the datasets, we compare models trained and tested on real graphs, on \pname synthetic traces, and models trained on \pname but evaluated on real graphs. provenance domain training on ProvCreator retains 65-70\% of the real-data F1, demonstrating that the synthetic graphs preserve class signal. When transferred to real traces the performance drops further, revealing a domain gap but still indicating non-trivial generalization from synthetic to real data.
\begin{figure}[h!]
    \centering
    \begin{subfigure}{0.3\textwidth}
        \centering
        \resizebox{\textwidth}{!}{%
            \includegraphics{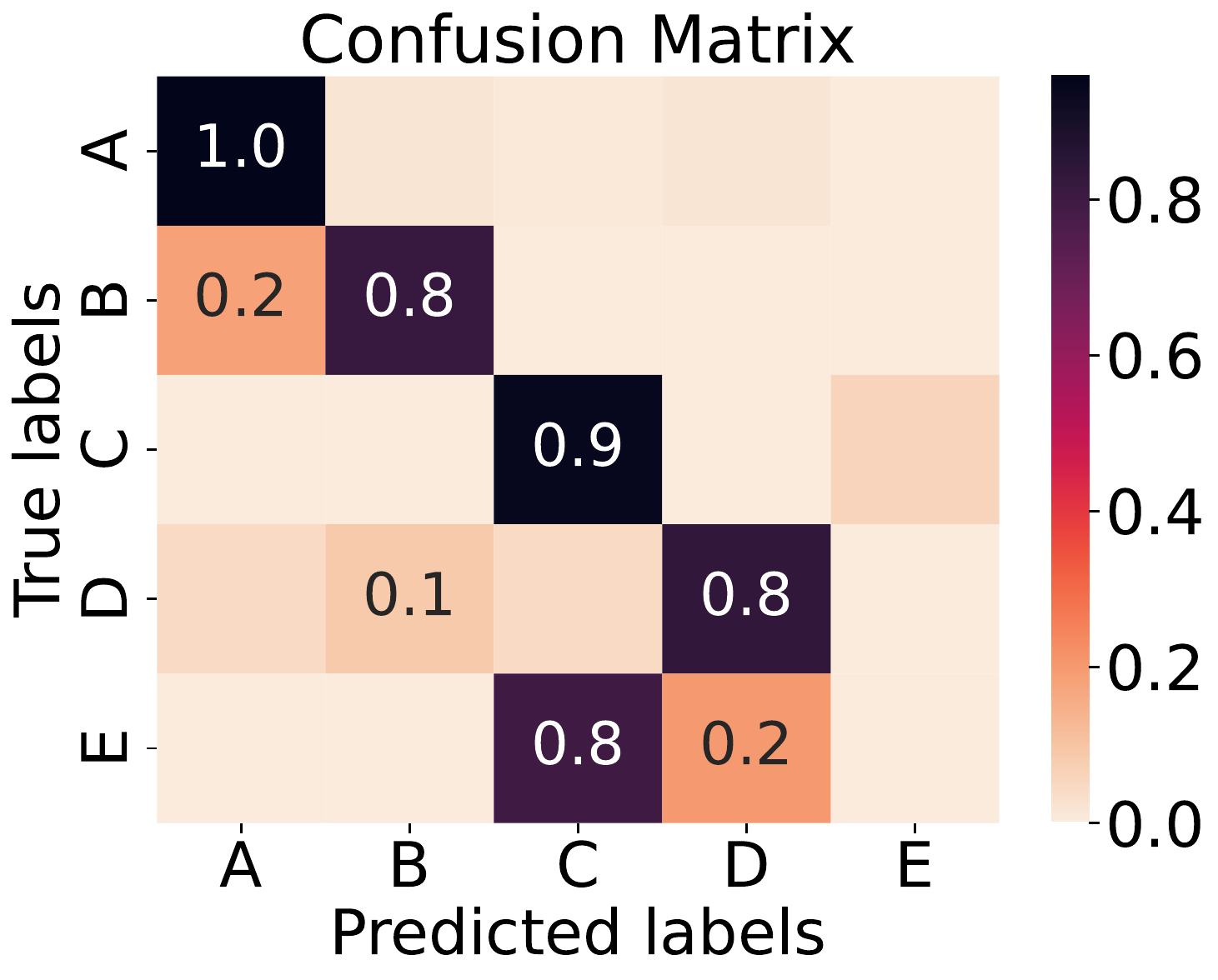}
        }
        \caption{Real \code{firefox.exe}.}
        \label{fig:miss-real-firefox}
    \end{subfigure}
    \hfill
    \begin{subfigure}{0.3\textwidth}
        \centering
        \resizebox{\textwidth}{!}{%
            \includegraphics{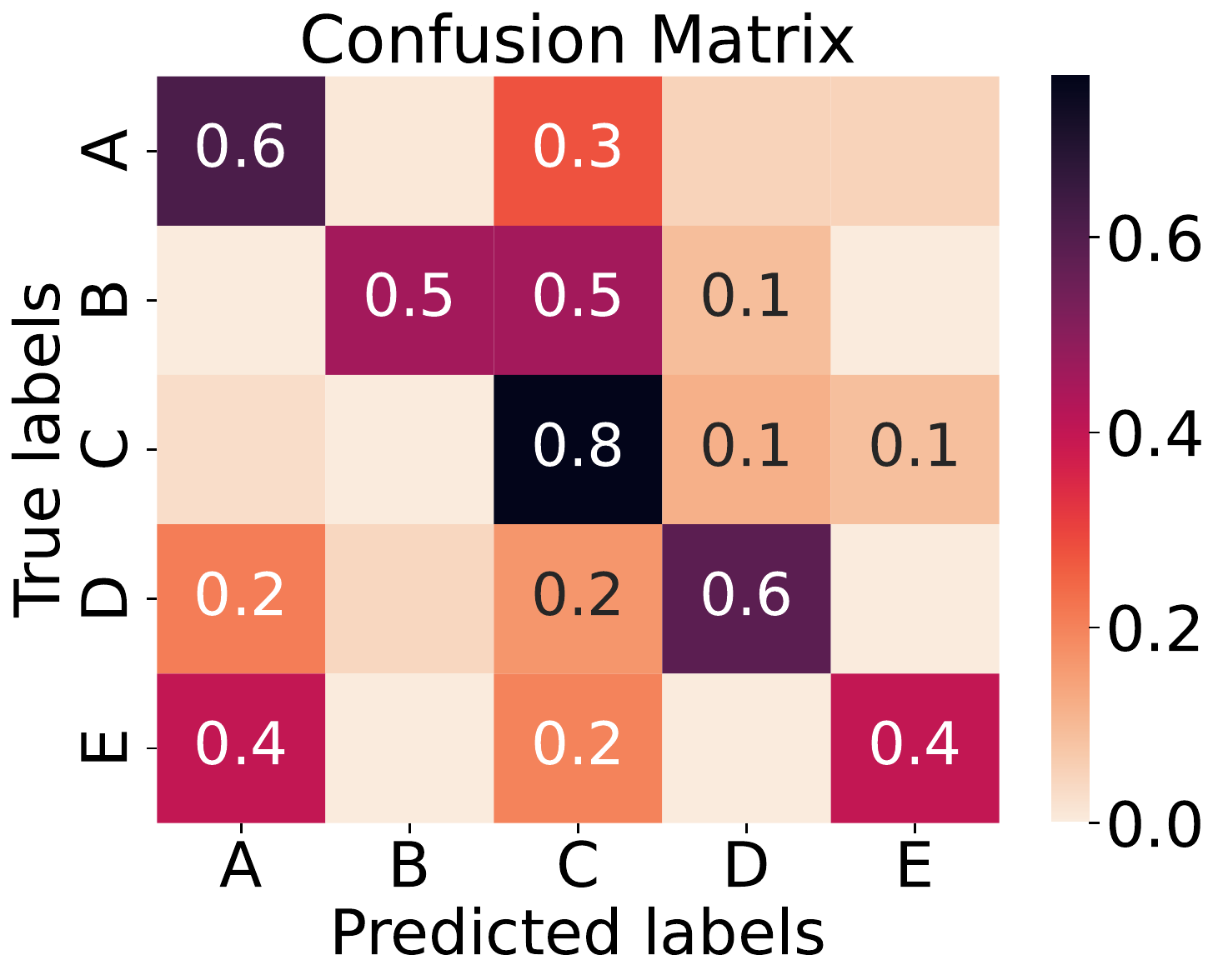}
        }
        \caption{\pname \code{firefox.exe}.}
        \label{fig:miss-gdss-firefox}
    \end{subfigure}
    \hfill
    \begin{subfigure}{0.36\textwidth}
        \centering
        \resizebox{0.8333\textwidth}{!}{%
            \includegraphics{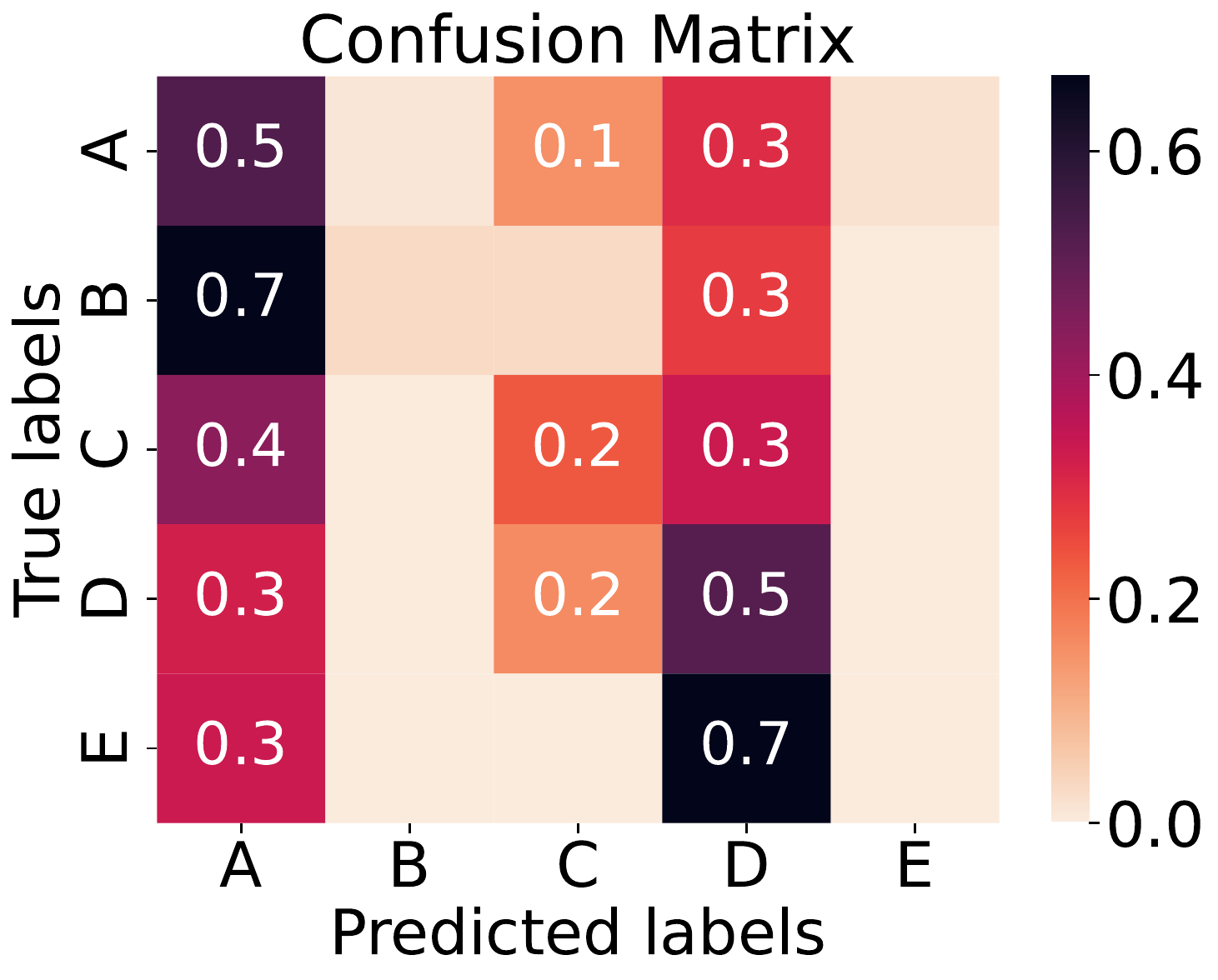}
        }
        \caption{Train on \pname but test on real \code{firefox.exe}.}
        \label{fig:miss-syn-firefox}
    \end{subfigure}
    %%%%%
    \begin{subfigure}{0.3\textwidth}
        \centering
        \resizebox{\textwidth}{!}{%
            \includegraphics{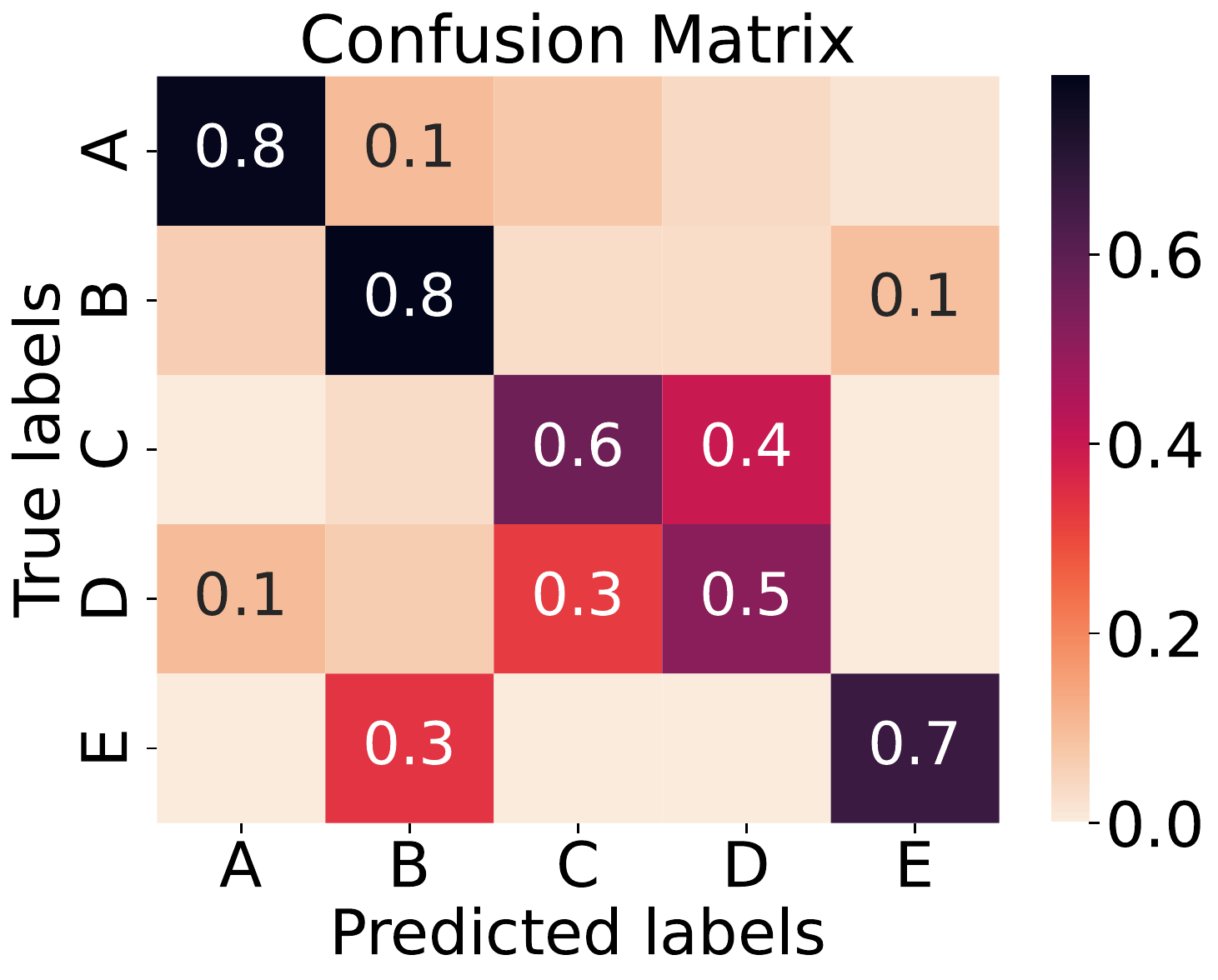}
        }
        \caption{Real \code{powershell.exe}.}
        \label{fig:miss-real-powershell}
    \end{subfigure}
    \hfill
    \begin{subfigure}{0.3\textwidth}
        \centering
        \resizebox{\textwidth}{!}{%
            \includegraphics{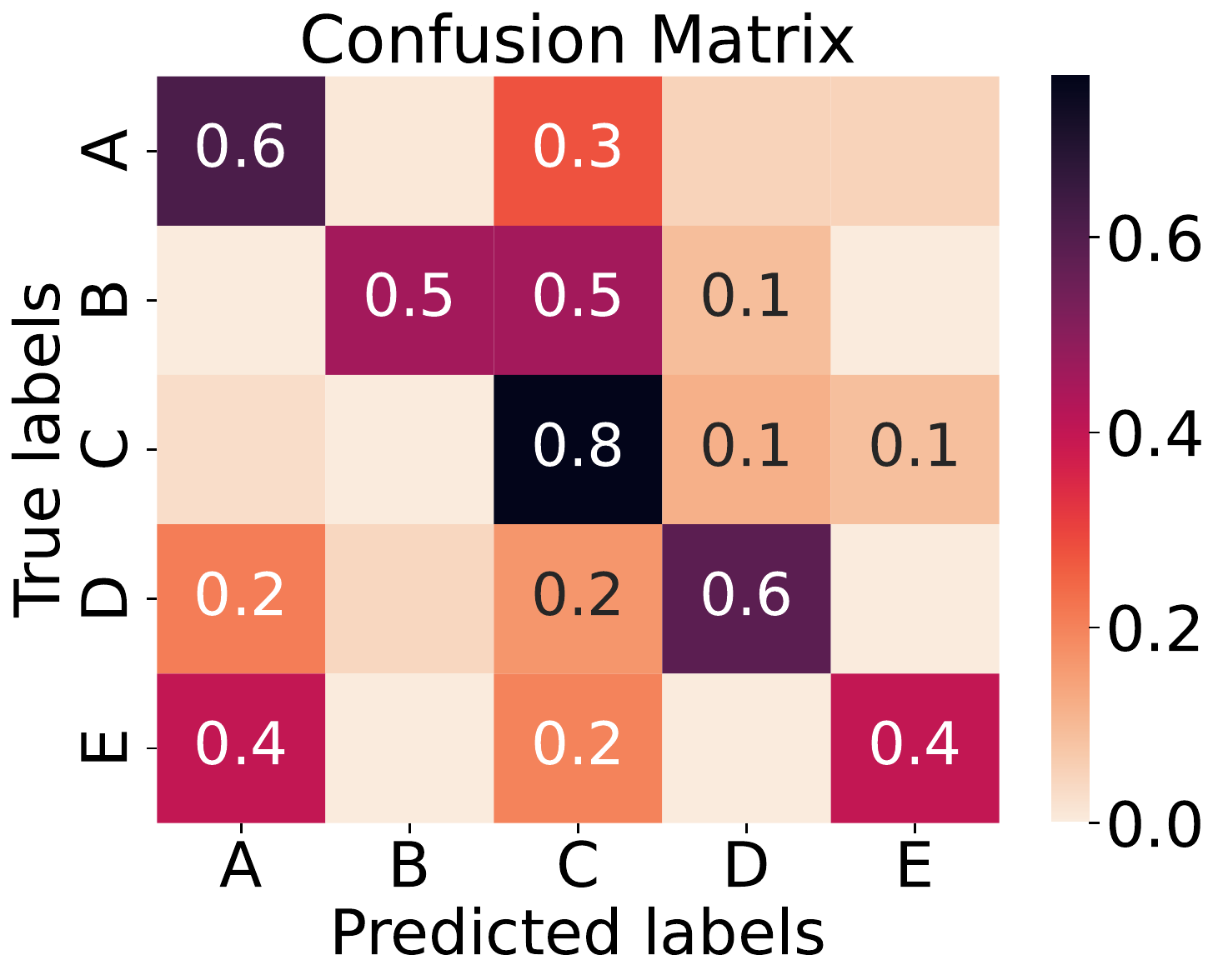}
        }
        \caption{\pname \code{powershell.exe}.}
        \label{fig:miss-gdss-powershell}
    \end{subfigure}
    \hfill
    \begin{subfigure}{0.36\textwidth}
        \centering
        \resizebox{0.8333\textwidth}{!}{%
            \includegraphics{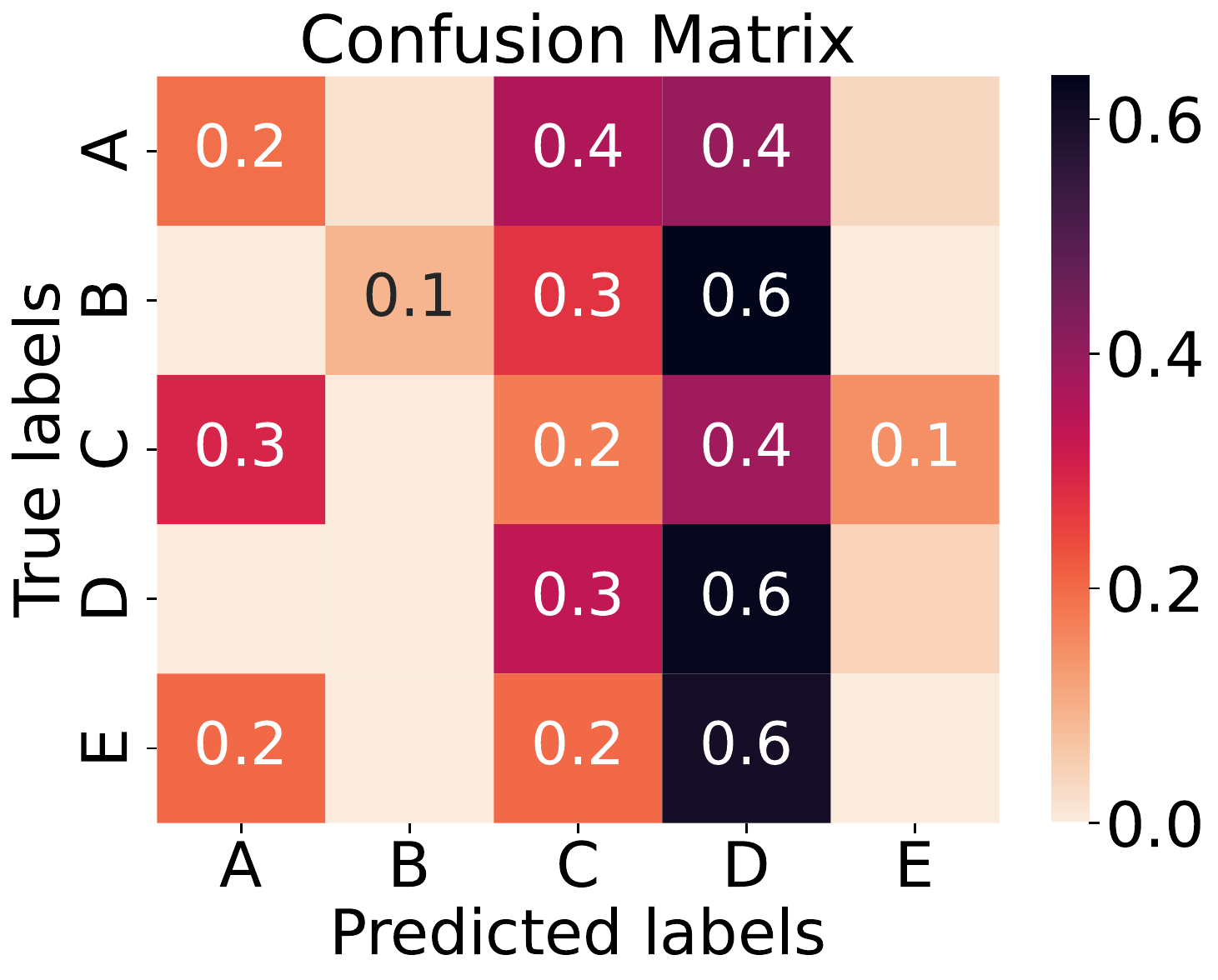}
        }
        \caption{Train on \pname but test on real \code{powershell.exe}.}
        \label{fig:miss-syn-powershell}
    \end{subfigure}

    \caption{Normalized confusion matrices for subprogram classification task.}
    \label{fig:confusion-matrix}
\end{figure}

\heading{Program Classification.}
Program classification is a supervised learning problem in which a model must infer the program executed by a host process from its provenance footprint. Concretely, given a provenance graph rooted at a \code{firefox.exe} or \code{powershell.exe} process, the objective is to predict the issued command by analyzing the system resources referenced in the graph. The complete set of class labels is listed in \autoref{sec:label}. In subplot (a,d) in \autoref{fig:confusion-matrix}, we see that the classifier trained and tested on real graph; (b,e) the classifier is trained and tested on \pname synthetic graphs; (c,f) the classifier trained on \pname graphs but evaluated on held-out real traces.

\autoref{fig:confusion-matrix} reveal that \pname preserves the class-discriminative structure observed in real data. When models trained on \pname  are transferred to real traces (c,f), accuracy declines but remains substantial, demonstrating that the synthetic dataset is realistic to support cross-domain generalization and mitigate real-data scarcity.

These gains hold for both the \code{firefox.exe} and \code{powershell.exe} scenarios, underscoring \pname's ability to generate representative synthetic graphs that bolster classification performance where training data are sparse.

%! root=../main.tex
\section{Conclusion}

In this paper, we introduced \pname, a novel synthetic graph generation framework designed for heterogeneous graphs with nonuniform text attributes.
By jointly considering the graph structure, node attributes, and program class labels, \pname addresses program class imbalance in system provenance datasets with synthetic graphs that contain rich textual attributes. Further, the inclusion of flexible node attribute indicators enables the generation of semantically nuanced textual attributes, with multiple attributes per node.
Our evaluation on real-world programs demonstrates improved graph structural fidelity compared to prior works, as well as improved utility in security-relevant downstream tasks.
\pname{} marks a step towards applying synthetic data generation to support accuracy-critical applications that require rich graph attributes.

\heading{Reproducibility.}

% paper link: https://openreview.net/pdf?id=nO344avRib
All experimental code related to \pname is available at \url{https://anonymous.4open.science/r/provcreator-aio-4F83}. % Detailed insights regarding the experiments, encompassing dataset and model specifics, are available in Section x. For intricate details like hyper parameter search, consult Appendix x. In addition, the reproduced dataset for each baseline is in Appendix x.

\bibliographystyle{iclr2024_conference}
\bibliography{refs/iclr2024_conference.bib}

%! root=../main.tex
\appendix
\section{Appendix}

\subsection{Graph Tokenization}

\autoref{alg:tokenization} presents the procedure for graph tokenization. It processes a heterogeneous graph $G(V,E)$, consisting of a node set $V$ and an edge set $E$, along with node types $T$, edge types $E_T$, and attributes linked to both nodes and edges.

\begin{algorithm}[h]
    \caption{Graph Tokenization}
    \label{alg:tokenization}
    \begin{algorithmic}
        \State \textbf{Input:} Graph $G(V,E)$, node types $T$, edge types $E_T$
        \State Initialize token sequence $S = [\text{\textless bog\textgreater}]$
        \State Initialize node set $N = \{\}$
        \For{each edge $(u,v) \in E$}
            \State Append \textless boe\textgreater{} to $S$
            \State Append \textless node$_u$\textgreater{} to $S$
            \State Append \textless node$_v$\textgreater{} to $S$
            \State Append the edge type token for $(u,v)$ to $S$
            \State Append \textless bof\textgreater{} to $S$
            \For{each attribute in the edge}
                \State Append the attribute token to $S$
            \EndFor
            \State Append \textless eof\textgreater{} to $S$
            \If{$u$ is not in $N$}
                \State Add $u$ to $N$
                \State Append \textless bon\textgreater{} to $S$
                \State Append \textless node$_u$\textgreater{} to $S$
                \State Append \textless ntype$_{T[u]}$\textgreater{} to $S$
                \For{each attribute in the node}
                    \State Append \textless bof\textgreater{} to $S$
                    \State Append the attribute token to $S$
                    \State Append \textless eof\textgreater{} to $S$
                \EndFor
                \State Append \textless eon\textgreater{} to $S$
            \EndIf
            \If{$v$ is not in $N$}
                \State Add $v$ to $N$
                \State Append \textless bon\textgreater{} to $S$
                \State Append \textless node$_v$\textgreater{} to $S$
                \State Append \textless ntype$_{T[v]}$\textgreater{} to $S$
                \For{each attribute in the node}
                    \State Append \textless bof\textgreater{} to $S$
                    \State Append the attribute token to $S$
                    \State Append \textless eof\textgreater{} to $S$
                \EndFor
                \State Append \textless eon\textgreater{} to $S$
            \EndIf
        \EndFor
        \State Append \textless eog\textgreater{} to $S$uence $S$
    \end{algorithmic}
\end{algorithm}

\subsection{Dataset Statistics}
\label{sec:dataset-statistics}

\begin{figure}[H]
    \centering
    \begin{subfigure}{0.3\textwidth}
        \centering
        \includegraphics[width=\textwidth]{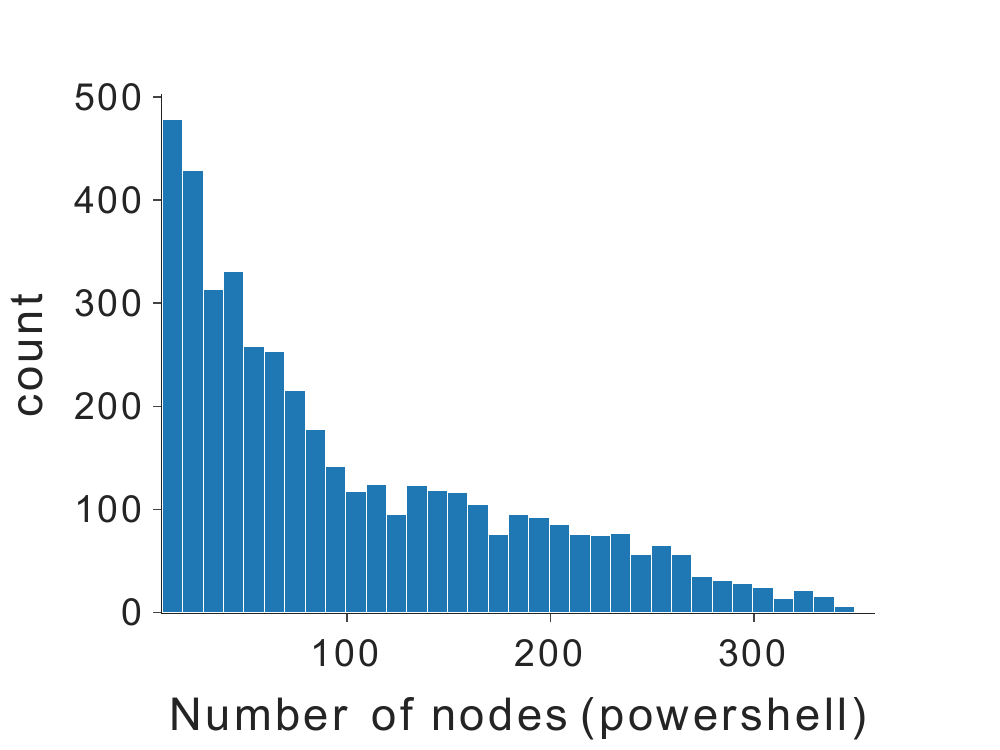}
        \caption{}
        \label{fig:powershell-num-nodes}
    \end{subfigure}
    \hfill
    \begin{subfigure}{0.3\textwidth}
        \centering
        \includegraphics[width=\textwidth]{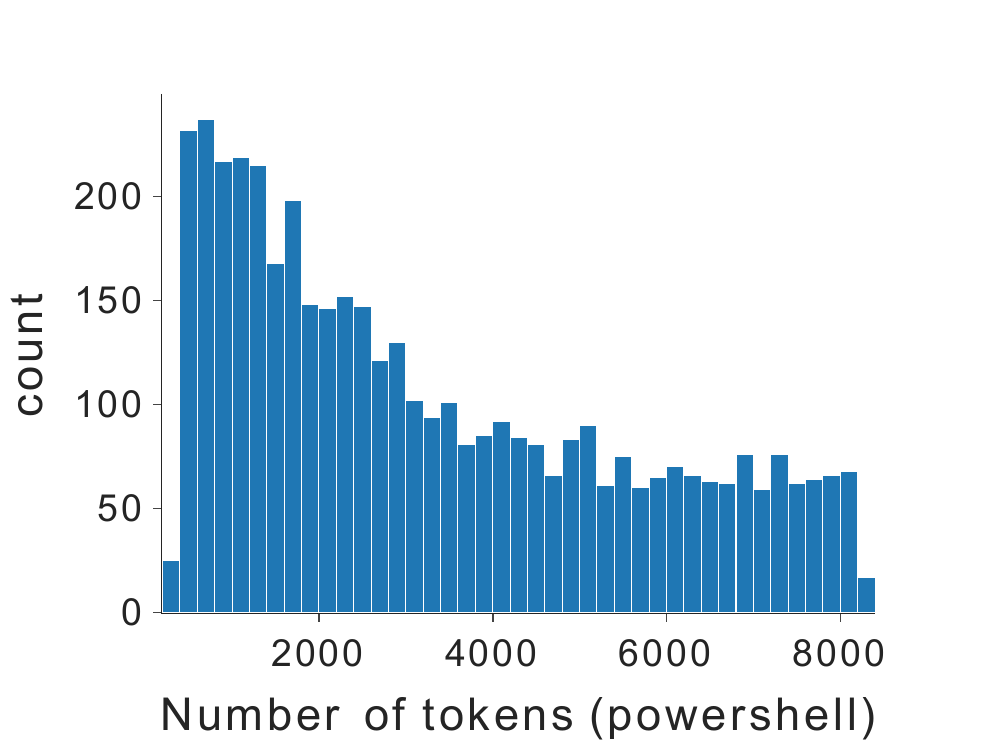}
        \caption{}
        \label{fig:powershell-num-tokens}
    \end{subfigure}
    \hfill
    \begin{subfigure}{0.3\textwidth}
        \centering
        \includegraphics[width=\textwidth]{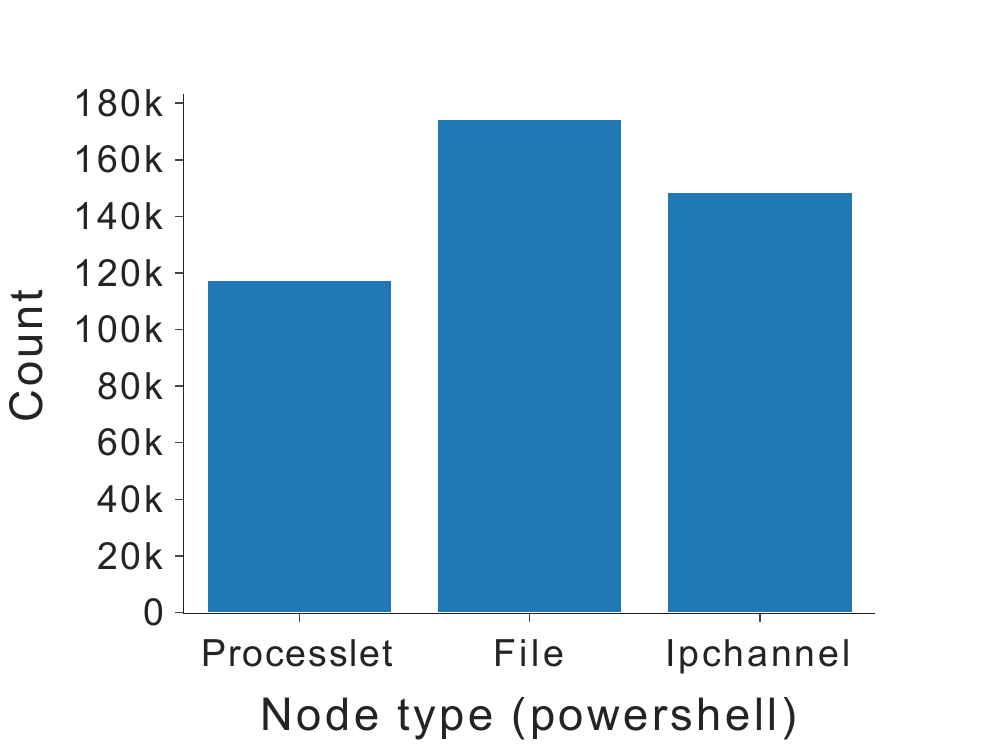}
        \caption{}
        \label{fig:powershell-node-types}
    \end{subfigure}
    \caption{Dataset statistics for PowerShell graphs.}
    \label{fig:powershell-asi-stats}
\end{figure}

\begin{figure}[H]
    \centering
    \begin{subfigure}{0.3\textwidth}
        \centering
        \includegraphics[width=\textwidth]{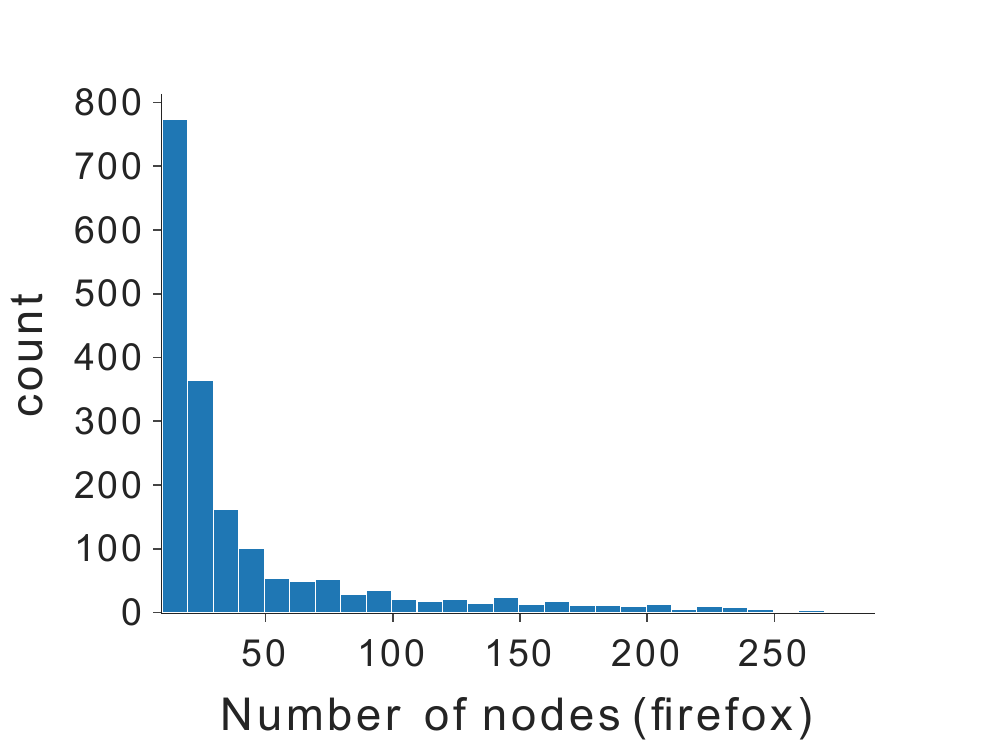}
        \caption{}
        \label{fig:firefox-num-nodes}
    \end{subfigure}
    \hfill
    \begin{subfigure}{0.3\textwidth}
        \centering
        \includegraphics[width=\textwidth]{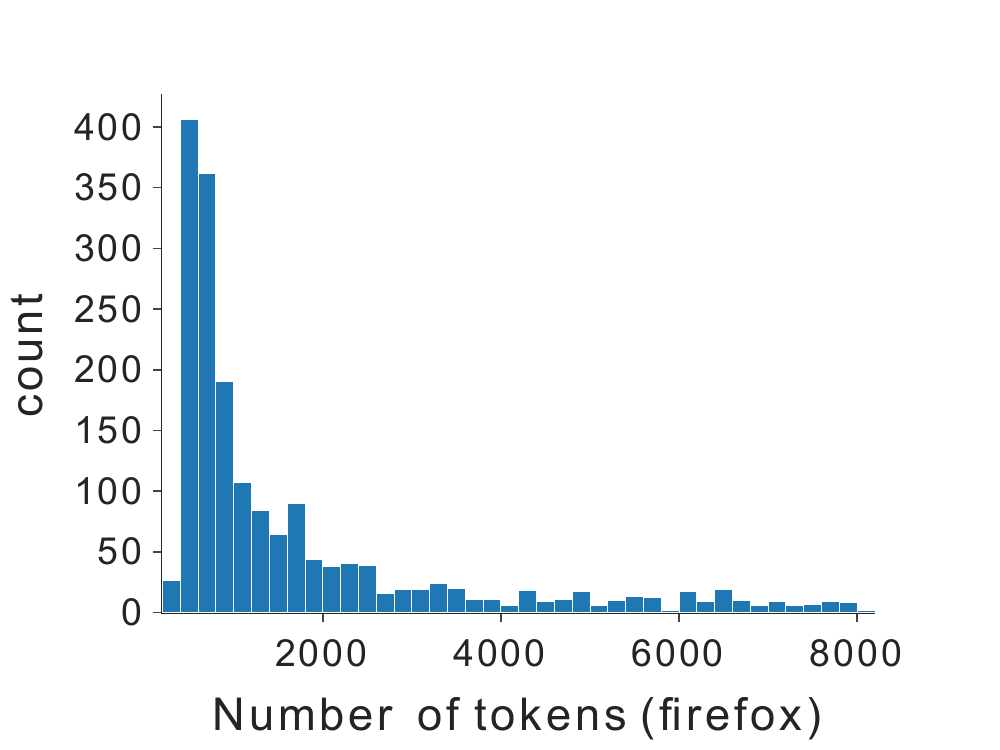}
        \caption{}
        \label{fig:firefox-num-tokens}
    \end{subfigure}
    \hfill
    \begin{subfigure}{0.3\textwidth}
        \centering
        \includegraphics[width=\textwidth]{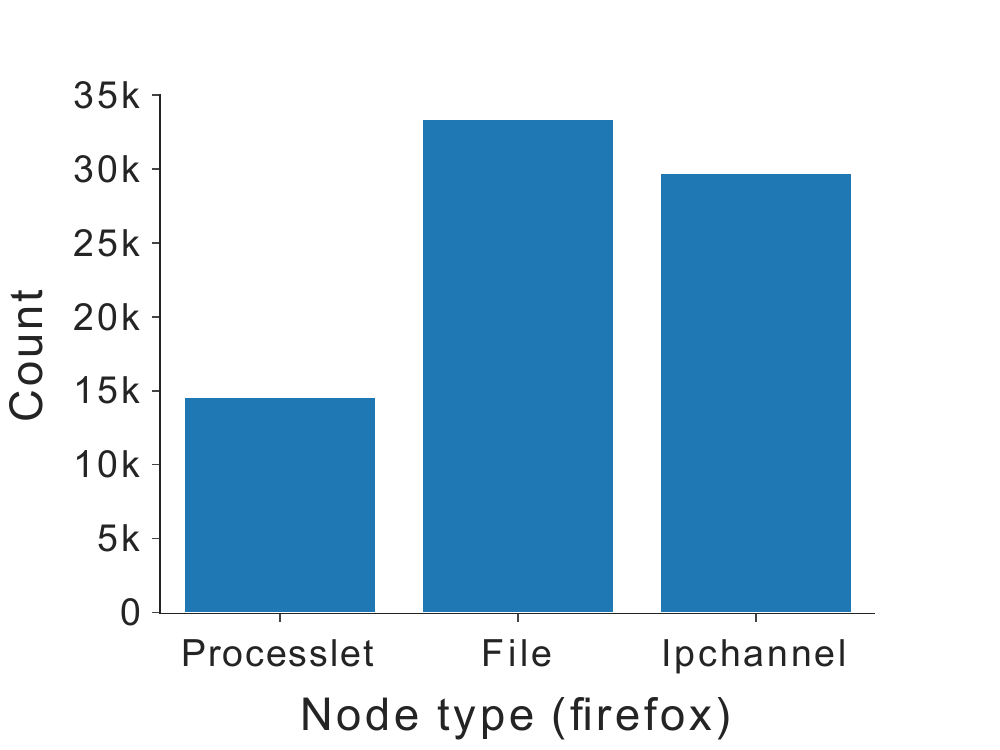}
        \caption{}
        \label{fig:firefox-node-types}
    \end{subfigure}
    \caption{Dataset statistics for Firefox graphs.}
    \label{fig:firefox-asi-stats}
\end{figure}

Figures \ref{fig:powershell-asi-stats}, and \ref{fig:firefox-asi-stats} show the dataset statistics for the provenance datasets used in training the generative models. Figure (a) shows the distribution of node counts. Figure (b) shows the number of tokens after tokenization. Figure (c) shows the distribution of node types.

\subsection{Computational Cost for Training and Generation}
\label{sec:computational-cost}
The training and generation times for \pname are shown in Table \ref{tab:training-time}. The training time is the time taken to train the model on the training dataset, while the generation time is the time taken to generate a single graph on average. The training time is measured in hours, while the generation time is measured in seconds. All experiments were conducted on a machine with 2x Intel Gold 6226R CPUs and 8x NVIDIA A6000 GPUs. During training, each model uses 4 GPUs and we set the per\_device\_train\_batch\_size to $4$, allowing us to run multiple training jobs in parallel. During generation, we only use 2 GPUs. 

\begin{table}[H]
    \centering
    \caption{Training and generation times for \pname.}
    \label{tab:training-time}
    \begin{tabular}{lcc}
        \toprule
        \textbf{Dataset} & \textbf{Training Time (hours)} & \textbf{Generation Time (seconds)} \\
        \midrule
        Firefox & 12 & 145 \\
        PowerShell & 15 & 106 \\
        \bottomrule
    \end{tabular}
\end{table}

\subsection{Validation of Generated Attributes}
\label{sec:validation-attributes}
We uses a list of regular expressions shown in \autoref{tab:attrib-validation-regex} to validate the generated attributes. These rules are designed to capture the expected format and constraints of each attribute type.

\begin{table}[H]
    \centering
    \caption{Validation rules for generated attributes.}
    \label{tab:attrib-validation-regex}
    \small
    \begin{tabular}{p{0.18\textwidth}p{0.76\textwidth}}
        \toprule
        \textbf{Attribute} & \textbf{Regular Expression (python format)}\\
        \toprule
        Executable Name (Windows) & \verb@^[A-Za-z]:\\(?:[^\\/:*?\"<>|\r\n]+\\)*[^\\/:*?\"<>|\r\n]*\.exe$@ \\
        \midrule
        Executable Name (Linux) & \verb@^/([^/\0]+/)*[^/\0]+$@ \\
        \midrule
        File Path (Windows) & \verb@^[A-Za-z]:\\(?:[^\\/:*?\"<>|\r\n]+\\)*[^\\/:*?\"<>|\r\n]*$@ \\
        \midrule
        File Path (Linux) & \verb@^/([^/\0]+/)*[^/\0]+$@ \\
        IP Address and Port & \verb@^(?:\d{1,3}\.){3}\d{1,3}\|\d{1,5}$@ \\
        \bottomrule
    \end{tabular}%

\end{table}

\subsection{Chosen subprograms of \code{firefox.exe} and \code{powershell.exe}}\label{sec:label}

We chose \code{firefox.exe} and \code{powershell.exe} programs because they are popular targets for impersonation by advanced cyber threats \citep{survivialism2021sp}, and their behavior is largely determined by their command-line arguments, which provides clear class labels.

\begin{table}[H]
    \centering
    \caption{Label descriptions for \texttt{firefox.exe}.}
    \label{tab:firefox-label}
    \resizebox{1.\columnwidth}{!}{%
    \begin{tabular}{cp{9cm}p{9cm}}
        \toprule
\textbf{Label} & \textbf{Command} & \textbf{Description} \\
\midrule
A & \texttt{firefox.exe --MOZ\_LOG ...} & Sets the logging level and other log-related options for specified components of Firefox. \\
\midrule
B & \texttt{firefox.exe --backgroundtask ...} & Invokes Firefox in a headless background task mode. This is used for periodic maintenance tasks, like checking for updates or sending telemetry, and can run even when the main Firefox browser is not open. \\
\midrule
C & \texttt{firefox.exe} & This starts the main Firefox web browser process. \\
\midrule
D & \texttt{firefox.exe -contentproc ...} & This is a subprocess of a Firefox browser instance that handles web content in a particular tab. \\
\midrule
E & \texttt{firefox.exe -osint ...} & This command line argument is used when Firefox is invoked by the operating system's shell to handle a search query or open a link. \\
\bottomrule
    \end{tabular}%
    }
\end{table}
\begin{table}[H]
    \centering
    \caption{Label descriptions for \code{powershell.exe}.}
    \label{tab:powershell-label}
    \resizebox{1.\columnwidth}{!}{%
    \begin{tabular}{cp{9cm}p{9cm}}
        \toprule
\textbf{Label} & \textbf{Command} & \textbf{Description} \\
\midrule
A & \texttt{powershell.exe -ExecutionPolicy AllSigned -NoProfile -NonInteractive -Command "\& \{\textbf{Inline Script}\}"} & Executes an inline PowerShell script used by Windows Defender to verify the hash of scripts it uses. \\
\midrule
B & \texttt{powershell.exe -Nologo -Noninteractive -NoProfile -ExecutionPolicy Bypass; Get-DeliveryOptimizationStatus ...} & Retrieves the status of Windows Update delivery optimization jobs and filters them to identify jobs from a local cache source and meeting specific criteria including a minimum file size. \\
\midrule
C & \texttt{powershell.exe -ExecutionPolicy Bypass -NoProfile -Command "Add-Type '\textbf{C\# Code 1}'}; \textbf{\dots} & Adds a .NET type, then calls a function from that type. These PowerShell command is found in a Windows Defender script. \\
\midrule
D & \texttt{powershell.exe -ExecutionPolicy Bypass -NoProfile -Command "Add-Type '\textbf{C\# Code 2}'}; \textbf{\dots} & Adds a .NET type (different from class C), then calls a function from that type. This PowerShell command is found in a Windows Defender script. \\
\midrule
E & \texttt{powershell.exe -NoLogo -Noninteractive -NoProfile -ExecutionPolicy Bypass \& 'C:\textbackslash{}Windows\textbackslash{}CCM\textbackslash{}SystemTemp\textbackslash{}\textbf{GUID}'} & Executes a PowerShell script in a temporary directory used by Microsoft System Center Configuration Manager (SCCM). Likely for automated management, software deployment, or configuration tasks. \\
\bottomrule
        \end{tabular}%
    }
\end{table}

\subsection{GDSS Model Architecture}

\begin{figure}[H]
    \centering
    \includegraphics[width=0.8\textwidth]{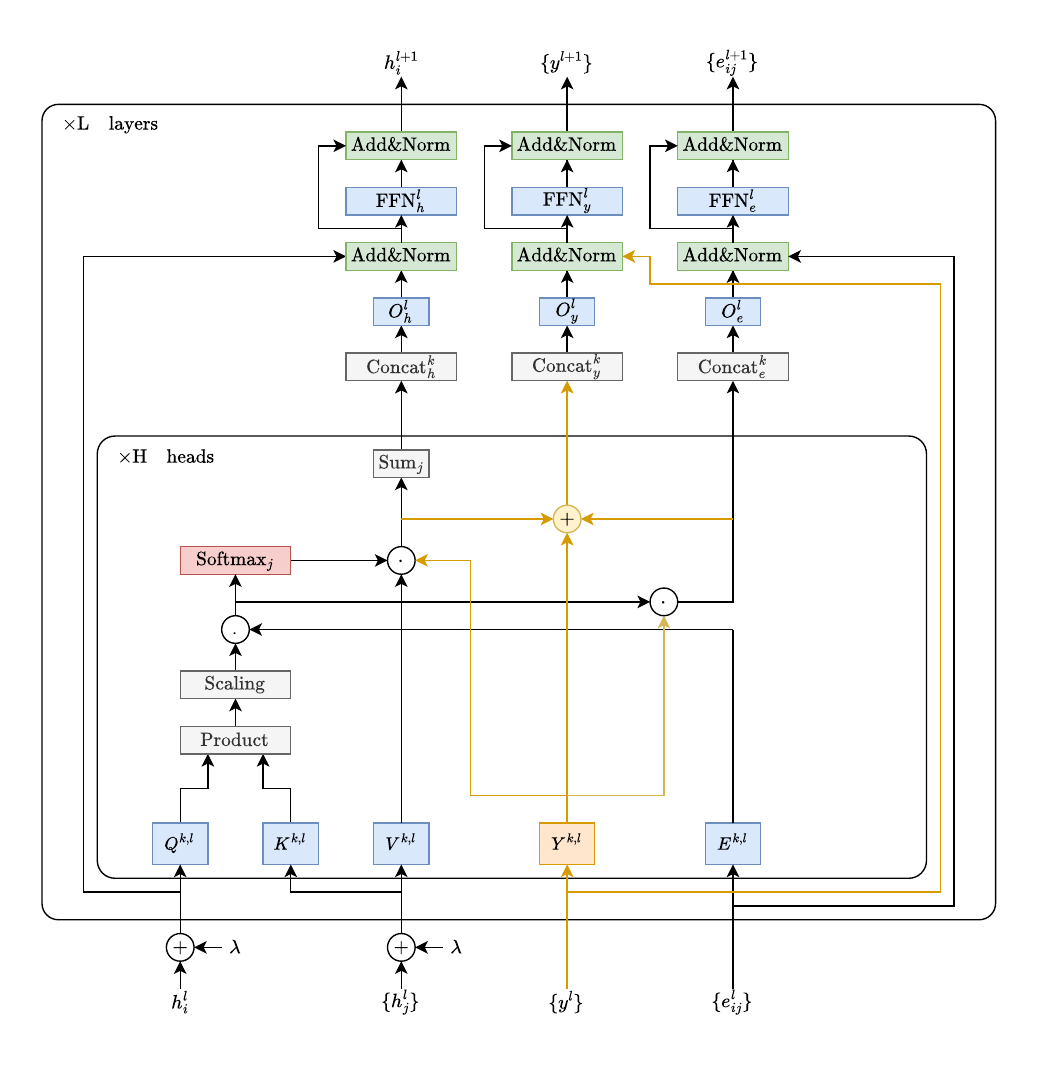}
    \caption{Structure Generation Model Architecture}
    \label{fig:structure-generation-model}
\end{figure}

Figure \ref{fig:structure-generation-model} shows the architecture of the model. In our experiments, we set $L=12$, $H=8$, the dimension of $h$ is $256$, the dimension of $e$ and $y$ is $128$. The orange arrow highlights the conditional generation components we introduced.

% Figure \ref{fig:attribute-generation-model} shows the architecture of the attribute generation model. In our implementation, we set $L=4$, $M=12$, hidden dimension of the GCN is $512$, Embedding size of the transformer is $512$ and max length is $128$. We used the RobertaTokenizer which gives us the token size of $50265$.

% \begin{figure}[H]
%     \centering
%     \includegraphics[width=0.6\textwidth]{figs/ProvCreatorAttribTransformer.drawio.pdf}
%     \caption{Attribute Generation Model Architecture}
%     \label{fig:attribute-generation-model}
% \end{figure}

\subsection{Example of Generated Graphs}

\begin{figure}[H]
    \centering
    \begin{subfigure}{0.45\textwidth}
        \centering
        \includegraphics[width=\textwidth]{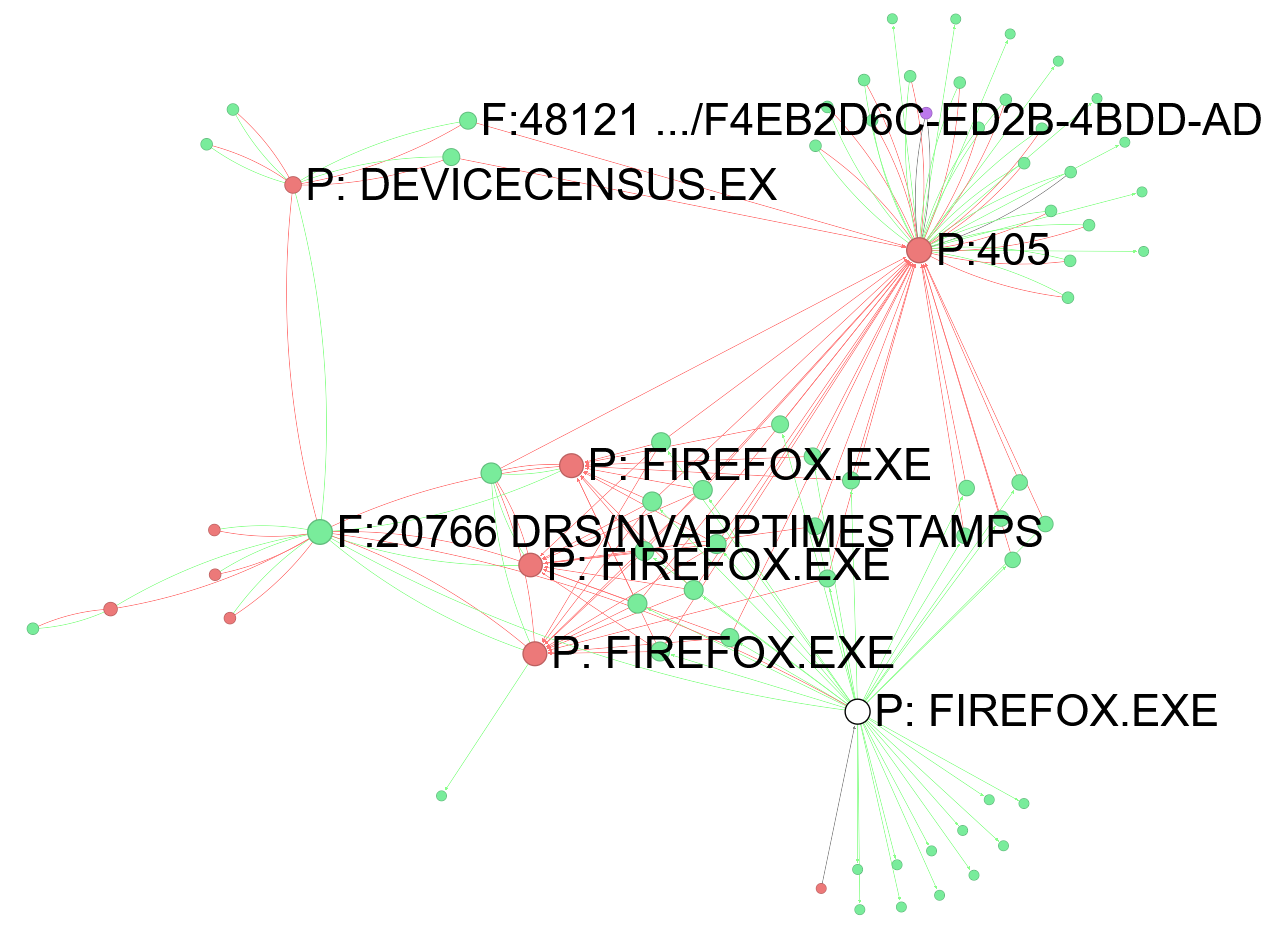}
        \caption{Real \code{firefox.exe} class D sample}
        \label{fig:real-example-1}
    \end{subfigure}
    \hfill
    \begin{subfigure}{0.45\textwidth}
        \centering
        \includegraphics[width=\textwidth]{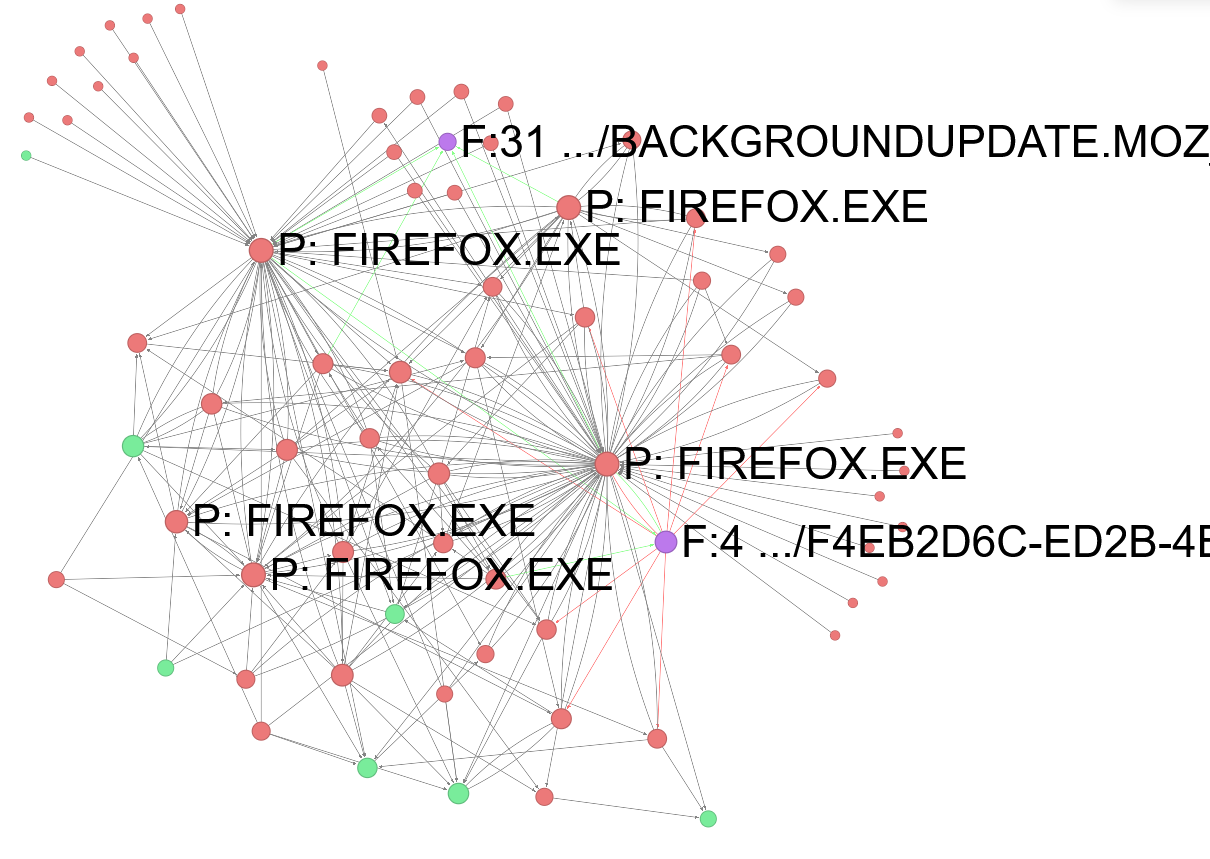}
        \caption{Synthetic \code{firefox.exe} class D sample}
        \label{fig:attribute-example-1}
    \end{subfigure}
    \vfill
    \begin{subfigure}{0.45\textwidth}
        \centering
        \includegraphics[width=\textwidth]{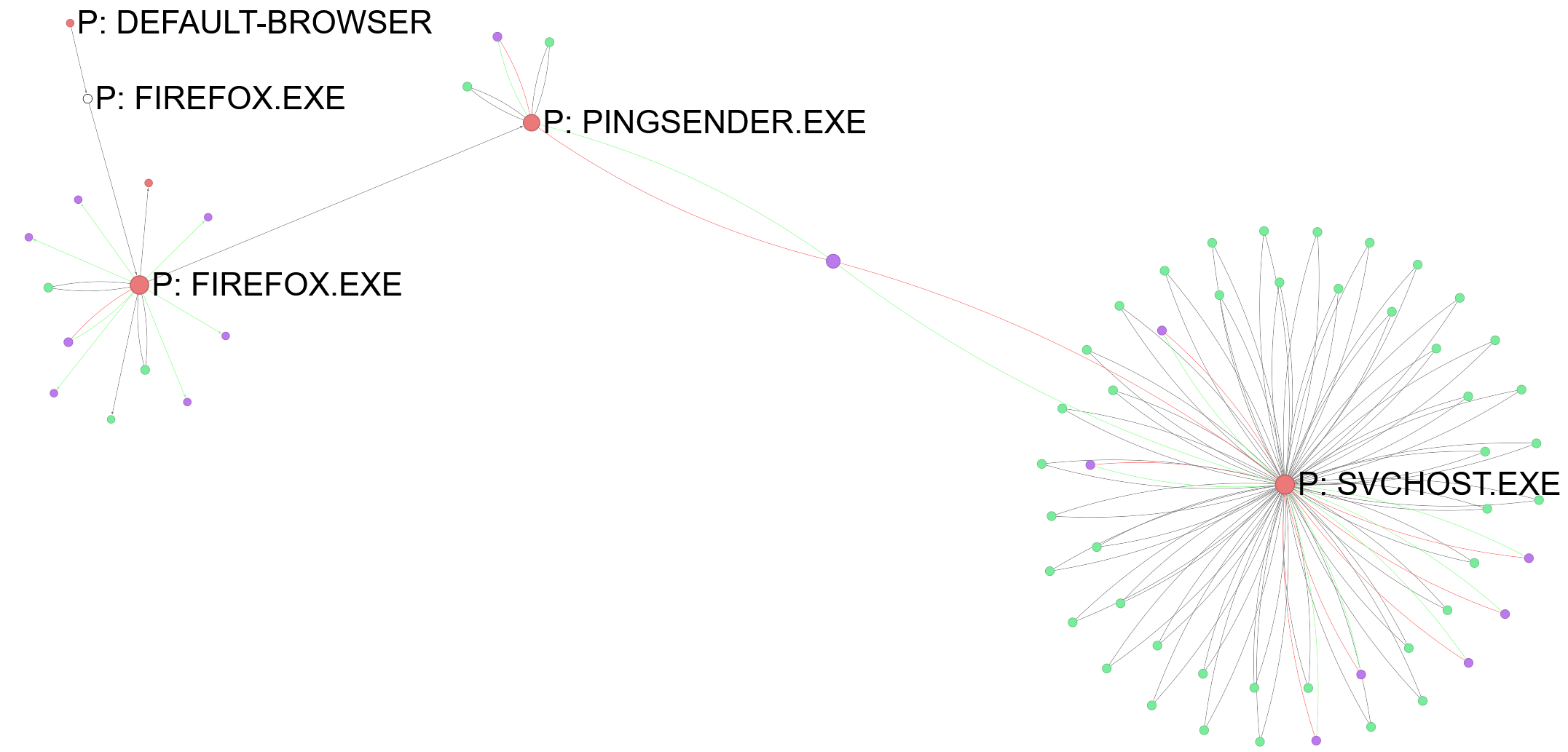}
        \caption{Real \code{firefox.exe} class B sample}
        \label{fig:real-example-2}
    \end{subfigure}
    \hfill
    \begin{subfigure}{0.45\textwidth}
        \centering
        \includegraphics[width=\textwidth]{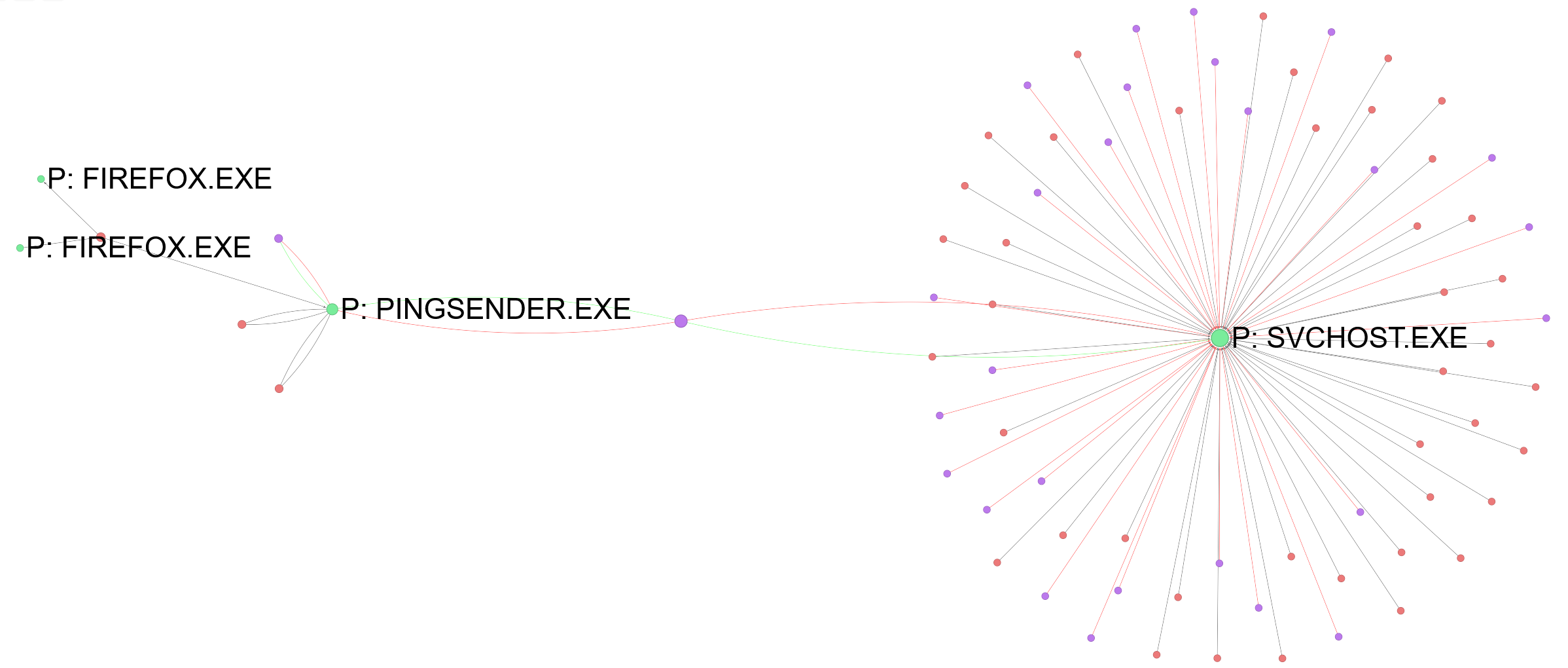}
        \caption{Synthetic \code{firefox.exe} class B sample}
        \label{fig:attribute-example-2}
    \end{subfigure}
    \vfill
    \begin{subfigure}{0.45\textwidth}
        \centering
        \includegraphics[width=\textwidth]{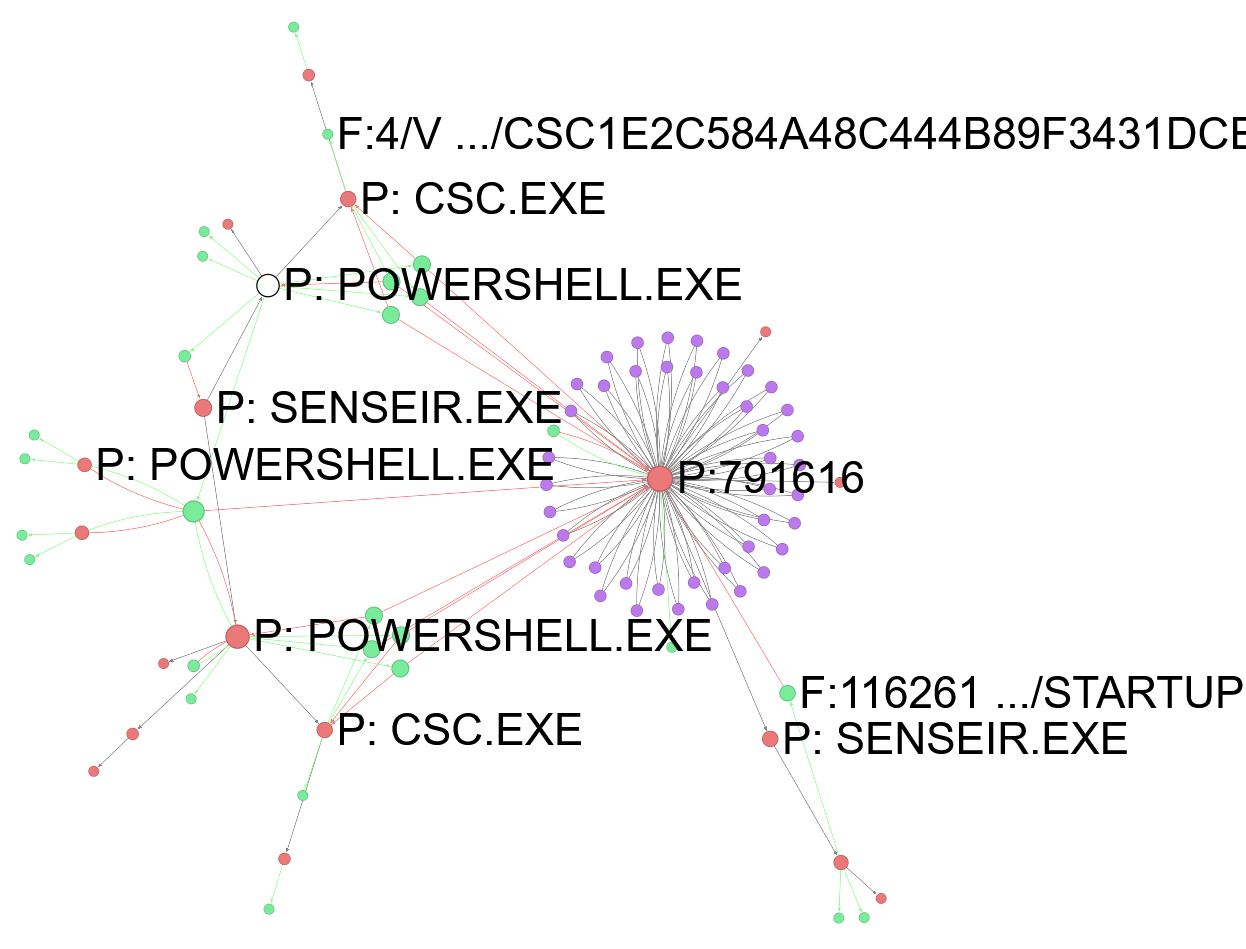}
        \caption{Real \code{powershell.exe} class D sample}
        \label{fig:structure-example-3}
    \end{subfigure}
    \hfill
    \begin{subfigure}{0.45\textwidth}
        \centering
        \includegraphics[width=\textwidth]{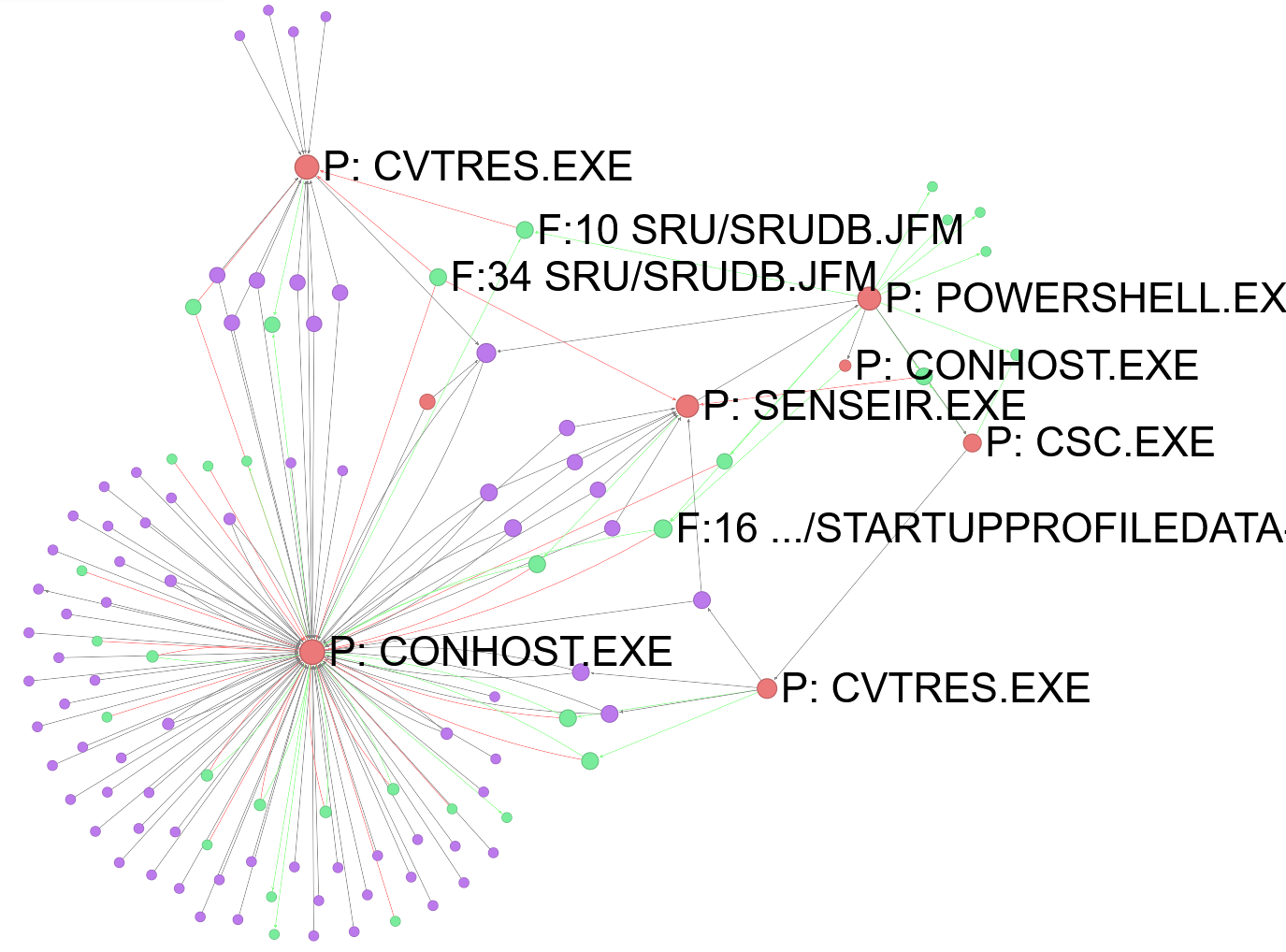}
        \caption{Synthetic \code{powershell.exe} class D sample}
        \label{fig:attribute-example-3}
    \end{subfigure}
    \caption{Examples of generated graphs}
    \label{fig:generated-graphs}
\end{figure}

Figure \ref{fig:generated-graphs} shows examples of real and generated graphs. The left column shows real provenance graphs from our training dataset, and the right column shows graphs from the same class generated by \pname.

\end{document}